\documentclass{article} 
\usepackage{iclr2027_conference,times}

\usepackage{amsmath,amsfonts,bm}

\def\eqref#1{equation~\ref{#1}}

\def\1{\bm{1}}

\DeclareMathAlphabet{\mathsfit}{\encodingdefault}{\sfdefault}{m}{sl}
\SetMathAlphabet{\mathsfit}{bold}{\encodingdefault}{\sfdefault}{bx}{n}

\usepackage{multirow}
\usepackage{hyperref}
\usepackage{url}
\usepackage{booktabs}
\usepackage[table]{xcolor}
\usepackage{threeparttable}
\usepackage{adjustbox}
\usepackage{xcolor}
\usepackage[most]{tcolorbox}
\usepackage{amssymb}
\usepackage{algorithm}
\usepackage{tabularx}
\usepackage{makecell}
\usepackage{enumitem}
\usepackage{algpseudocode}
\usepackage{subcaption}
\usepackage[most]{tcolorbox}
\usepackage{xcolor}
\usepackage{hyperref}

\definecolor{citeblue}{RGB}{45,85,145}

\hypersetup{
  colorlinks=true,
  citecolor=citeblue,
  linkcolor=citeblue,
  urlcolor=citeblue
}
\definecolor{qwenblue}{HTML}{5B7DB1}
\definecolor{glmviolet}{HTML}{7867A8}
\definecolor{kimigreen}{HTML}{5F8F7B}

\definecolor{boxgray}{HTML}{F7F8FA}
\definecolor{textgray}{HTML}{777777}
\definecolor{linegray}{HTML}{D9DCE2}

\tcbuselibrary{listings,breakable,skins}

\definecolor{promptframe}{HTML}{3B6EA8}
\definecolor{promptback}{HTML}{F5F8FC}

\title{Beyond Token Alignment: Event Completion for Cross-Tokenizer On-Policy Distillation}

\author{
Jiacheng Liu\textsuperscript{1}\thanks{Equal contribution.},
Jingwei Song\textsuperscript{1}\footnotemark[1],
Qituan Zhang\textsuperscript{2},
Siheng Chen\textsuperscript{1},
Linfeng Zhang\textsuperscript{1}\thanks{Corresponding author.}
\\[2mm]
\textsuperscript{1}Shanghai Jiao Tong University
\textsuperscript{2}Fudan University
}

\definecolor{OPDBlue}{HTML}{2F6BFF}
\definecolor{MetricGray}{HTML}{777777}

\usepackage{amsthm}

\definecolor{BestBlue}{RGB}{198, 222, 245}
\definecolor{SecondBlue}{RGB}{236, 244, 251}
\newcommand{\best}[1]{%
    \cellcolor{BestBlue}\textbf{#1}%
}

\newcommand{\snd}[1]{%
    \cellcolor{SecondBlue}\underline{#1}%
}

 \newtcblisting{promptbox}[1]{%
    breakable,
    enhanced,
    arc=2pt,
    boxrule=0.5pt,
    colframe=promptframe,
    colback=promptback,
    coltitle=white,
    fonttitle=\bfseries\small,
    title={#1},
    top=4pt,
    bottom=4pt,
    left=6pt,
    right=6pt,
    listing only,
    listing options={
      basicstyle=\ttfamily\small,
      breaklines=true,
      breakatwhitespace=true,
      keepspaces=true,
      columns=fullflexible,
      breakindent=0pt,
      postbreak={}
    }
  }

\usepackage[most]{tcolorbox}

\tcbset{
    reasoningcase/.style={
        enhanced,
        colback=gray!5,
        colframe=gray!40,
        coltext=black!85,
        coltitle=black,
        fonttitle=\bfseries\small,
        fontupper=\small,
        boxrule=0.5pt,
        arc=1.5mm,
        outer arc=1.5mm,
        left=3mm,
        right=3mm,
        top=2.5mm,
        bottom=2.5mm,
        toptitle=2.2mm,
        bottomtitle=2.2mm,
        boxsep=0pt
    }
}

\iclrfinalcopy 
\begin{document}

\maketitle
\lhead{Preprint.} 

\vspace{-4mm}
\begin{abstract}
\vspace{-3mm}
On-policy distillation (OPD) transfers knowledge between language models through teacher supervision on student-generated trajectories. With different tokenizers, a single teacher token may require multiple student tokens to generate, creating intermediate states where the event is entered but not yet completed. Existing cross-tokenizer methods align tokens or text spans to construct comparable prediction targets. We study a complementary problem after partial generation: once the student produces a prefix of a teacher token, multiple next tokens may complete the same remaining bytes, but the teacher only specifies the required completion rather than how probability should be divided among these valid continuations. We introduce \textbf{Event-Set Completion Distillation (ESCD)}, which complements cross-tokenizer probability alignment with completion-set supervision. ESCD aggregates prefix-related teacher events and supervises the \textbf{total probability of byte-compatible one-step student completions}, avoiding tokenizer-dependent probability splits among individual tokens. The method reuses student trajectories and predictions, requiring neither additional rollouts nor changes to the student vocabulary. Experiments demonstrate consistent gains in mathematics, code, and scientific reasoning across model families and tokenizers, extending to large-scale MoE distillation from a \textbf{1T teacher to a 35B student}. Local analyses show that retaining completion sets better matches the reference supervision, while one-step completion covers over 99\% of observed compatible teacher mass after partial event entry in the studied tokenizer pairs. These findings support event entry and event completion as complementary supervision targets for cross-tokenizer knowledge transfer. Code will be released on GitHub.

\begin{figure}[!htp]
    \centering
    \includegraphics[width=\linewidth]{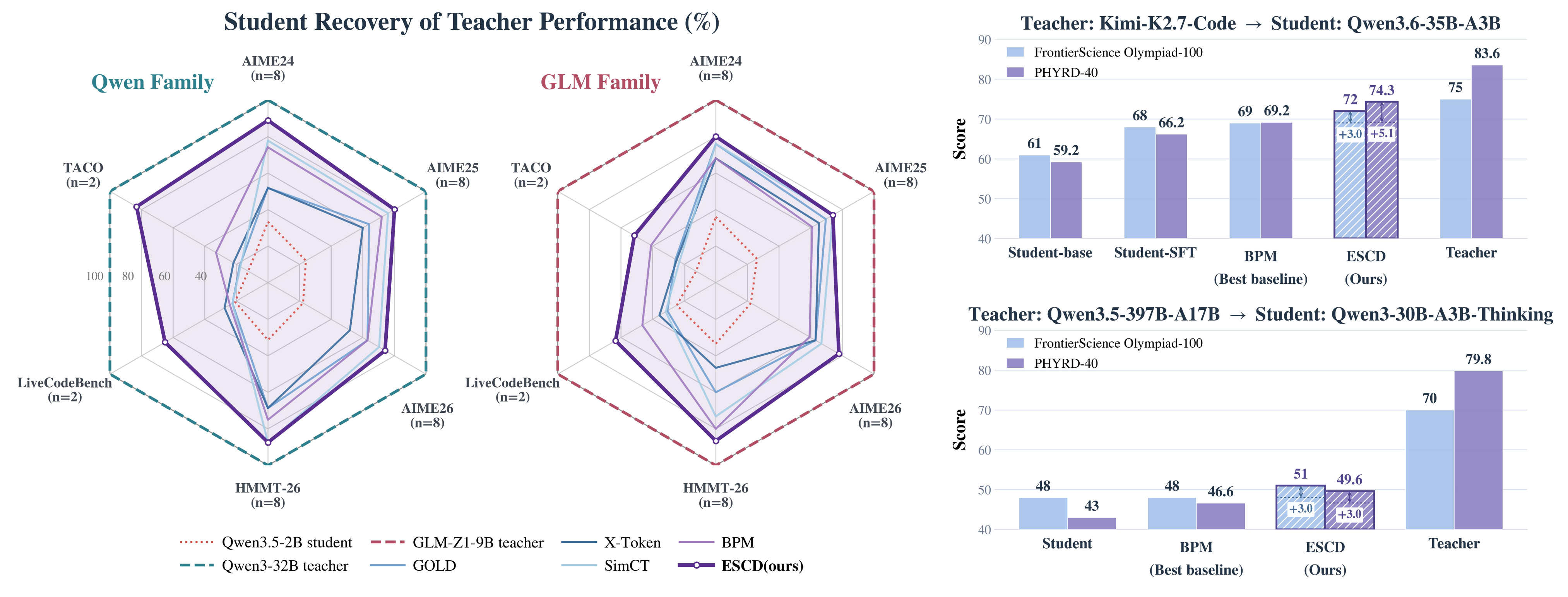}

\caption{\textbf{Cross-tokenizer distillation results across model scales.} \textbf{Left:} Teacher-normalized pass@$n$ scores on six mathematics and code benchmarks, with Qwen3.5-2B distilled from Qwen3-32B or GLM-Z1-9B. Recovery is the Student score divided by the corresponding teacher score, expressed as a percentage; 100\% indicates matching teacher performance. We use $n=8$ for mathematics and $n=2$ for code. \textbf{Right:} Absolute scores on FrontierScience Olympiad and PHYRD-40 for two large-scale heterogeneous MoE pairs. Together, these results show consistent improvements across task domains, model families, and scales.}
\label{fig:showcase}
\end{figure}

\end{abstract}
\clearpage
\section{Introduction}
\label{sec:introduction}

\begin{figure*}[hbt]
    \centering
    \includegraphics[width=\textwidth]{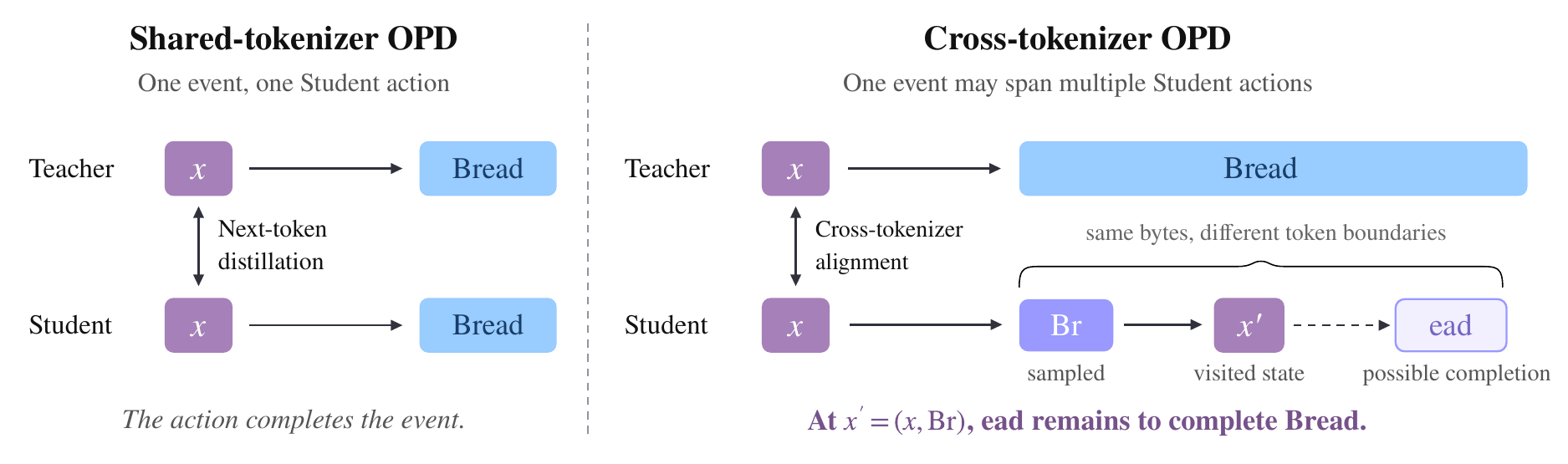}
    \vspace{-6mm}
    \caption{
    \textbf{On-policy distillation with shared and different tokenizers.}
    \textbf{Left:} a shared tokenizer allows direct next-token supervision.
    \textbf{Right:} a teacher token such as \texttt{Bread} may span multiple student actions.
    After sampling \texttt{Br}, the student reaches $x'=(x,\texttt{Br})$ with residual bytes \texttt{ead} still required to complete the event.
    The dashed arrow indicates a possible continuation, highlighting the distinction between entering an event and completing it.
    }
    \label{fig:intro}
\end{figure*}

The diversity of large language models creates opportunities for knowledge transfer across model families, architectures, and scales. A capable teacher and a smaller, deployment-efficient student may differ in model design, training data, and specialization. Knowledge distillation provides a framework for transferring these capabilities through teacher supervision~\citep{hinton2015distillingknowledgeneuralnetwork}. On-policy distillation (OPD) is well suited to autoregressive generation because it obtains teacher feedback on student-generated trajectories~\citep{lin2020autoregressiveknowledgedistillationimitation,gu2026minillmonpolicydistillationlarge,agarwal2024onpolicydistillationlanguagemodels}. Training under the student-induced state distribution reduces the mismatch between training contexts and generation states, including prefixes the teacher would not typically generate independently.

Many OPD formulations assume a shared tokenizer and therefore a common next-token prediction space. This assumption becomes restrictive in cross-family distillation, where both vocabularies and token boundaries can differ substantially. The mismatch concerns not only which token represents a piece of text, but also how many generation steps are needed to produce it. Existing cross-tokenizer methods establish comparable supervision through token mappings, distribution matching, and text- or byte-level alignment~\citep{uld,simct,bpm}. SimCT compares aligned continuation scores, while BPM also constructs conditional token targets inside a teacher token. Both therefore incorporate predictions beyond the first student action. We focus on a complementary question: \textit{after a student action partially realizes a teacher event, how should supervision constrain the total probability of valid completions at the resulting state?}

To make this question concrete, consider the teacher token \texttt{Bread}, which the student may realize as \texttt{['Br', 'ead']} (Figure~\ref{fig:intro}). After sampling \texttt{Br}, the rollout moves from $x$ to $x'=(x,\texttt{Br})$, leaving the residual bytes \texttt{ead}. At $x'$, the residual specifies which continuations complete the event, but this constraint alone does not determine their relative probabilities. Importantly, the teacher event defines a byte-level completion constraint rather than a tokenizer-dependent decomposition into individual student tokens. Therefore, assigning probabilities among multiple valid completions introduces an additional allocation choice that is determined by the student tokenizer rather than by the teacher supervision itself. Existing token-level objectives can supervise this state while prescribing such distributions over individual completions. We investigate whether supervising the total probability of valid completions provides an effective alternative for cross-tokenizer transfer. This frames the \textbf{event-completion gap} as a target-design problem: how to supervise the remaining byte constraint while leaving the allocation within its completion set unspecified.

More generally, let $b_T$ and $b_S$ denote token byte realizations, and consider a teacher content token $v$ without a single-token student counterpart. If the student samples a content token $a$ whose bytes are a strict prefix of $b_T(v)$, the bytes decompose into the generated prefix and residual bytes $r$:

\begin{equation}
b_T(v)=b_S(a)\mathbin{\Vert}r,
\qquad
x'=(x,a),
\label{eq:intro-event-residual}
\end{equation}

where $\mathbin{\Vert}$ denotes byte-string concatenation. Once the prefix is generated, the remaining requirement concerns the continuation from $x'$. Its valid one-step completions form a set of native student tokens:

\begin{equation}
\mathcal C(r)=\left\{u\in V_S^{\mathrm{cont}}:r\preceq b_S(u)\right\},
\label{eq:intro-completion-set}
\end{equation}

where $V_S^{\mathrm{cont}}$ is the student content-token vocabulary, excluding special tokens, and $\preceq$ denotes byte-prefix inclusion. A valid token may complete the residual exactly or extend beyond its boundary: for residual \texttt{ead}, both \texttt{ead} and \texttt{eads} qualify if present in the vocabulary. These tokens satisfy the same residual byte constraint without necessarily representing equivalent full continuations. Matching separate token targets can additionally constrain their relative probabilities. This motivates a completion-set objective that supervises their aggregate probability while leaving within-set allocation to the remaining training objective.

We introduce \textbf{Event-Set Completion Distillation (ESCD)}, which combines byte-aligned root projection with explicit residual-event completion supervision. ESCD first aggregates prefix-related teacher events into groups, using the shortest member as a shared prefix constraint and summing their teacher probability mass. This aggregation preserves the included mass while coarsening longer members' constraints to the representative prefix. When a student action partially realizes this constraint, ESCD identifies the remaining bytes and constructs the valid one-step completion set at the naturally visited child state. For nonempty sets, ESCD applies a teacher-mass-weighted negative log-likelihood to their aggregate student probability, using probabilities from the native full-vocabulary distribution. The child objective supervises this aggregate probability without prescribing target probabilities for individual valid completions.
The on-policy trajectory provides both the conditioning state and the student predictions needed by this objective. Completion candidates are evaluated through their probabilities at the visited child; they do not require separate sampled continuations. The auxiliary loss can therefore provide a completion signal even when the student's subsequently sampled token falls outside the valid set. ESCD requires neither counterfactual rollouts nor modifications to the student vocabulary. Its current formulation is local: residual events without a valid one-step completion contribute no auxiliary loss, while root-level supervision is retained.

Empirically, ESCD improves performance across model families, task domains, and scales. In controlled experiments with Qwen3.5-2B~\citep{qwen3.5} as the student and Qwen3-32B~\citep{qwen3} or GLM-Z1-9B~\citep{glm2024chatglm,zai2025glmz1} as the teacher, ESCD matches or exceeds the strongest compared baseline on every reported metric. Gains reach $10.0$ percentage points in mathematical reasoning and are particularly pronounced in code generation: under the Qwen teacher, pass@2 increases from $18.1\%$ to $42.9\%$ on LiveCodeBench and from $18.7\%$ to $47.3\%$ on TACO. Relative to the best baseline on each benchmark, \textbf{teacher performance recovery} rises from $62.0\%$ to $79.8\%$ for Qwen and from $64.0\%$ to $72.2\%$ for GLM.
We further evaluate ESCD in large-scale MoE distillation with 397B Qwen and 1T Kimi teachers, transferring their capabilities to 30B and 35B students, respectively~\citep{kimi2.7,qwen3.6-35b-a3b}. ESCD improves scientific reasoning performance in both settings, with gains of $3.0$--$5.1$ points over the strongest compared baseline. These results support cross-tokenizer OPD as a practical approach to transferring capabilities from large MoE teachers to smaller students for downstream deployment.

Our main contributions are summarized as follows:

\begin{itemize}

\item \textbf{From event entry to event completion.}
We identify the \textbf{event-completion gap} in cross-tokenizer OPD: a student action can enter a teacher byte event without completing it, leaving a residual byte constraint at the visited student state. We show that completion should be separated from tokenizer-dependent probability allocation among individual student tokens, motivating completion-set supervision.

\item \textbf{Event-Set Completion Distillation.}
We propose ESCD, which aggregates prefix-related teacher events and uses their probability mass to supervise the total probability of byte-compatible student completions. ESCD complements root-level alignment with visited-child completion supervision, without prescribing individual completion probabilities, requiring counterfactual rollouts, or modifying the student vocabulary.

\item \textbf{Cross-family transfer at scale.}
ESCD achieves consistent improvements over compared baselines across mathematics, code, and scientific reasoning. The gains extend to heterogeneous MoE distillation with teachers up to 1T parameters, improving scientific reasoning benchmarks by 3.0--5.1 points. These results show the applicability of cross-tokenizer OPD across model families, architectures, and scales.

\end{itemize}

\section{Related Work}
\label{sec:related-work}

\subsection{Distillation for Language Models}

Classical knowledge distillation transfers softened Teacher predictions to a smaller Student~\citep{hinton2015distillingknowledgeneuralnetwork}, while SeqKD extends this approach to autoregressive generation through Teacher-decoded sequences~\citep{kim2016sequencelevelknowledgedistillation}. To reduce the mismatch between training and Student-induced states, ImitKD queries the Teacher on Student-generated prefixes~\citep{lin2020autoregressiveknowledgedistillationimitation}, MiniLLM optimizes reverse KL on Student samples~\citep{gu2026minillmonpolicydistillationlarge}, and GKD supports mixtures of Student- and data-generated sequences with alternative divergences~\citep{agarwal2024onpolicydistillationlanguagemodels}. Subsequent work explores generalized and skew divergences for sequence-level distillation~\citep{wen2023fdivergenceminimizationsequencelevelknowledge,ko2024distillmstreamlineddistillationlarge}. DistiLLM-2 further couples the objective with the source of training data through a contrastive formulation for Teacher- and Student-generated responses~\citep{ko2025distillm2}.

Recent studies investigate the reliability and efficiency of on-policy supervision. Entropy-Aware OPD augments reverse KL with forward KL at high-entropy Teacher predictions to preserve diversity~\citep{jin2026entropyaware}. Other work identifies compatibility between Teacher and Student thinking patterns as a factor in OPD success and proposes off-policy cold starts and Teacher-aligned prompt selection~\citep{li2026rethinkingopd}. Further analysis distinguishes rapid coverage of Student-visited states from the slower absorption of Teacher supervision~\citep{fu2026rethinkingopd2}. These findings motivate examining which states provide useful feedback and how feedback is translated into a training objective.

\subsection{Cross-Tokenizer Distillation}

Cross-tokenizer distillation must reconcile both vocabulary and sequence-boundary mismatches. FuseLLM approximately aligns heterogeneous token sequences using minimum edit distance~\citep{wan2024knowledgefusionlargelanguage}, while DSKD uses learned projections to compare predictions in compatible output spaces~\citep{zhang2024dualspaceknowledgedistillationlarge}. ULD matches sorted probability profiles without retaining token identity~\citep{uld}, and MultiLevelOT aligns logit distributions through token- and sequence-level optimal transport~\citep{cui2025multilevelot}. GOLD extends cross-tokenizer supervision to on-policy training by merging text-aligned spans and combining direct matching on shared tokens with rank-based matching on unmatched tokens~\citep{gold}.

Other approaches construct supervision through text or byte representations. ALM compares likelihoods of text-equivalent chunks~\citep{minixhofer2025universalcrosstokenizerdistillationapproximate}, CTLS develops cross-tokenization likelihood scoring algorithms~\citep{phan2026crosstokenizerlikelihoodscoringalgorithms}, and BLD introduces an auxiliary byte-level prediction interface~\citep{singh2026crosstokenizerllmdistillationbytelevel}. SimCT builds minimal aligned multi-token units and compares their continuation scores~\citep{simct}, while X-Token constructs sparse vocabulary projections from token correspondences and multi-token decompositions~\citep{sreenivas2026xtokenprojectionguidedcrosstokenizerknowledge}. BPM constructs Student-token targets through byte-prefix marginalization, including conditional targets at positions inside a teacher token~\citep{bpm}. More recently, ACTD addresses mapping noise through anchor supervision and regularization of the remaining Student vocabulary~\citep{zhang2026actd}.

\paragraph{Our contribution.}
Existing cross-tokenizer methods transfer supervision through aligned tokens, spans, or conditional distributions, but primarily construct comparable prediction targets rather than explicitly modeling completion after partial event entry. ESCD introduces \textbf{completion-set supervision} at naturally visited student child states, directly targeting the total probability of valid one-step completions while avoiding tokenizer-dependent probability allocation among individual completions. It aggregates prefix-related teacher events into representative prefix constraints and uses their combined probability mass to weight the completion loss. Combined with byte-aligned root projection, this objective couples \textbf{event entry and residual completion} without counterfactual rollouts. 
\vspace{-5mm}
\begin{figure*}[t]
    \centering
    \includegraphics[width=\textwidth]{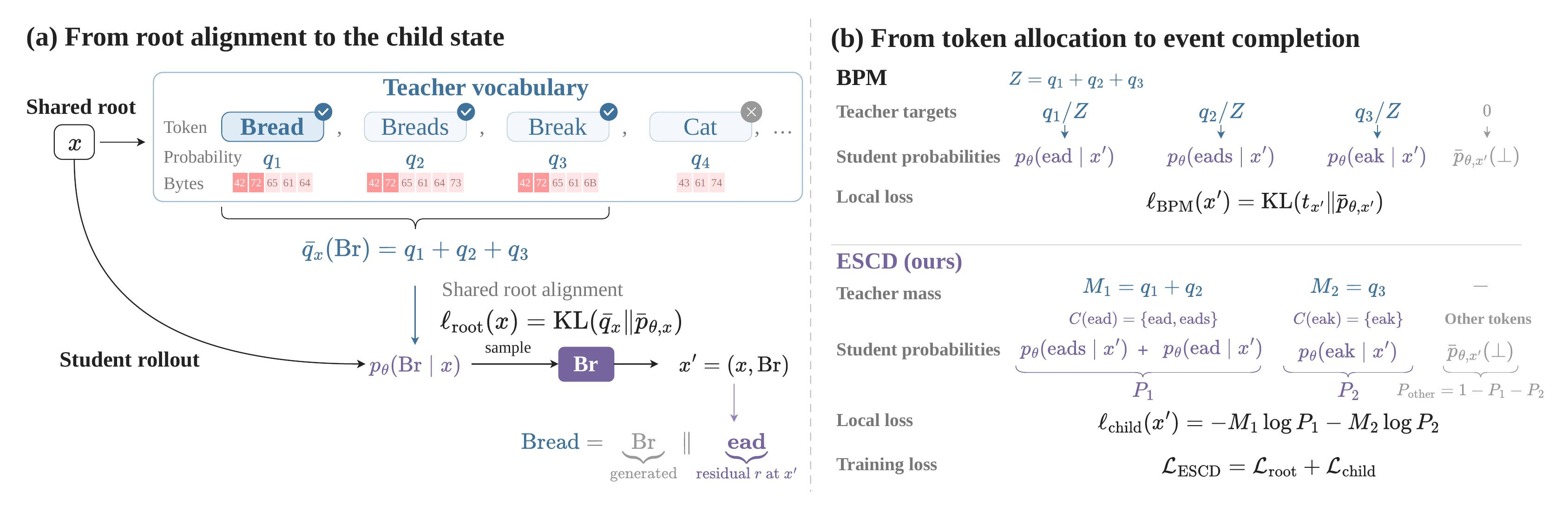}
    \vspace{-6mm}
\caption{
\textbf{ESCD: from event entry to event completion.}
\textbf{(a) From event entry to residual completion.}
At the shared root $x$, compatible teacher tokens contribute probability mass to the student action \texttt{Br}. Sampling \texttt{Br} reaches the visited state $x'=(x,\texttt{Br})$, where residual bytes such as \texttt{ead} remain to complete the teacher byte event \texttt{Bread}.
\textbf{(b) From token allocation to event completion.}
BPM constructs conditional targets for individual student tokens at the residual state, thereby specifying a probability allocation among possible completions. ESCD instead aggregates prefix-related teacher events and uses their combined mass to supervise the total probability of the byte-compatible one-step completion set. For example, both \texttt{ead} and \texttt{eads} satisfy the residual byte constraint \texttt{ead} without requiring a prescribed probability split. The final objective combines root alignment and completion supervision using student predictions at visited states, without additional rollouts.
}
    \label{fig:method}
\end{figure*}

\section{Method}
\label{sec:method}

We introduce \textbf{E}vent-\textbf{S}et \textbf{C}ompletion \textbf{D}istillation (\textbf{ESCD}). When a sampled student token begins but does not complete a teacher byte event, ESCD supervises the total probability of next tokens completing the remaining bytes. It combines this completion target with probability alignment at the preceding state. We define the sampled action's residual (Sec.~\ref{sec:event_completion}), then construct completion sets by aggregating prefix-related teacher events (Sec.~\ref{sec:completion_set}), and define the training objective (Sec.~\ref{sec:escd_objective}).

\subsection{From Event Entry to Event Completion}
\label{sec:event_completion}

Let $T$ be a frozen teacher and $S_\theta$ a trainable student with vocabularies $V_T$ and $V_S$. At a student-visited root context $x$, the aligned predictions correspond to the same response-byte prefix, with next-token distributions $q_T(\cdot\mid x)$ and $p_\theta(\cdot\mid x)$. Here, ``root'' denotes the starting state of a local event, not the beginning of the response. Each model represents this shared byte prefix using its own tokenizer. Let $b_T$ and $b_S$ map tokens to their exact byte strings. A teacher event specifies the bytes of a teacher token, while a student action is a sampled student token. Let $V_T^{\mathrm{cont}}$ and $V_S^{\mathrm{cont}}$ denote the corresponding content-token vocabularies.

A teacher token may span multiple student actions. Figure~\ref{fig:method}(a) illustrates how sampling \texttt{Br} at the root $x$ reaches the child state $x'=(x,\texttt{Br})$, leaving residual bytes \texttt{ead} for the teacher event \texttt{Bread}. Let $y=b_T(v)$ for a teacher content token $v\in V_T^{\mathrm{cont}}$. If the student samples a content token $a$ whose bytes form a nonempty strict prefix of $y$, the rollout reaches

\begin{equation}
x'=(x,a),
\qquad
r=y[|b_S(a)|:],
\label{eq:residual_event}
\end{equation}
where $b_S(a)\prec y$ denotes nonempty strict byte-prefix inclusion: $a$ produces part of $y$ but leaves the event incomplete. The residual $r$ is the suffix after removing $|b_S(a)|$ leading bytes from $y$.

Token-level alignment can supervise both event entry and intermediate student positions. In Figure~\ref{fig:method}(b), BPM assigns conditional target probabilities to individual student tokens, whereas ESCD supervises their total probability within each valid completion set. Thus, ESCD explicitly targets residual completion without requiring a particular probability split among valid next tokens. Appendix~\ref{sec:framework-existing-methods} compares the supervision targets, and Appendix~\ref{sec:framework-worked-example} gives a numerical example.

\subsection{Constructing Completion Sets}
\label{sec:completion_set}

The single-event example extends to multiple teacher candidates, whose prefix-related byte strings can impose overlapping completion requirements. ESCD groups these candidates under a shared representative prefix and uses their combined teacher mass to weight its completion target. We use the full teacher vocabulary as the candidate set, $K_T(x)=V_T$, and retain content tokens whose complete byte strings have no exact single-token counterpart in the student content vocabulary:

\begin{equation}
\mathcal V(x)
=
\left\{
v\in K_T(x)\cap V_T^{\mathrm{cont}}
:
\nexists u\in V_S^{\mathrm{cont}},\ b_T(v)=b_S(u)
\right\}.
\label{eq:escd-eligible-candidates}
\end{equation}

We connect two candidates in $\mathcal V(x)$ whenever either byte string is a prefix of the other and use the resulting connected components as aggregation groups. For component $g$ with members $V_g$, define
\begin{equation}
M_g=\sum_{v\in V_g}q_T(v\mid x),
\qquad
y_g=b_T(v_g^\star),
\qquad
v_g^\star\in\operatorname*{arg\,min}_{v\in V_g}|b_T(v)|,
\label{eq:teacher_aggregation}
\end{equation}
with deterministic tie breaking. The shortest member defines a prefix shared by all group members. Aggregation preserves their total teacher mass while coarsening their byte constraints to this representative prefix. For example, grouping \texttt{Bread} and \texttt{Breads} retains their shared \texttt{Bread} requirement but does not require the additional \texttt{s}. Thus, $M_g$ weights completion of the shared prefix, rather than exact reconstruction of every original teacher event.

For each group satisfying $b_S(a)\prec y_g$, the sampled action $a$ partially realizes the representative event, leaving residual bytes $r_g=y_g[|b_S(a)|:]$. At the visited child state $x'=(x,a)$, we define the valid one-step completion set and its aggregate probability:
\begin{equation}
\mathcal C(r_g)=\{u\in V_S^{\mathrm{cont}}:r_g\preceq b_S(u)\},
\qquad
P_\theta(\mathcal C(r_g)\mid x')
=
\sum_{u\in\mathcal C(r_g)}p_\theta(u\mid x'),
\label{eq:completion_set}
\end{equation}
where $\preceq$ denotes byte-prefix inclusion, allowing equality. Probabilities come from the student's native full-vocabulary distribution, without renormalizing over content tokens or the completion set. The sums play different roles: teacher aggregation determines the representative constraint's weight $M_g$, while student marginalization measures the probability of satisfying it in one token.

A valid completion must begin with the entire residual but may extend beyond it. In Figure~\ref{fig:method}(b), the \texttt{Bread}/\texttt{Breads} group has weight $M_1=q_1+q_2$. After sampling \texttt{Br}, both \texttt{ead} and \texttt{eads} complete the representative residual \texttt{ead}, giving $P_1=p_\theta(\texttt{ead}\mid x')+p_\theta(\texttt{eads}\mid x')$ in the illustrative vocabulary. Satisfying this shared constraint does not imply equivalent full continuations. For a fixed action, distinct eligible groups have disjoint completion sets because neither representative prefix is a prefix of the other. Empty completion sets contribute no child loss.

\subsection{Training Objective}
\label{sec:escd_objective}

We reuse BPM's byte alignment and root projection, without retaining its full interior- and spanning-position supervision. At an aligned root $x$, teacher mass is mapped to selected student tokens: Figure~\ref{fig:method}(a), for example, maps \texttt{Bread}, \texttt{Breads}, and \texttt{Break} to \texttt{Br}. The projected distributions $\bar q_x$ and $\bar p_{\theta,x}$ share these token entries and a complement $\bot$ holding their respective remaining mass (Appendix~\ref{sec:framework-escd}). Without renormalizing over explicit entries, we apply forward KL:
\begin{equation}
\ell_{\mathrm{root}}(x)
=
\operatorname{KL}\!\left(
\bar q_x \,\middle\|\, \bar p_{\theta,x}
\right).
\label{eq:escd-root-loss}
\end{equation}

For the sampled action $a$, let $\mathcal G(x,a)$ contain groups with $b_S(a)\prec y_g$ and nonempty completion sets $\mathcal C(r_g)$, where $r_g=y_g[|b_S(a)|:]$. At the visited child state $x'=(x,a)$, the completion loss is
\begin{equation}
\ell_{\mathrm{child}}(x,a)
=
-\sum_{g\in\mathcal G(x,a)}
M_g
\log\left[
\sum_{u\in\mathcal C(r_g)}
p_\theta(u\mid x,a)
\right].
\label{eq:child_objective}
\end{equation}

The weights $M_g$ retain their root teacher mass without renormalization over eligible groups; student probabilities use the native full-vocabulary distribution at $x'$. This loss supervises completion-set probability without individual token targets, regardless of whether the next sampled token belongs to the set. It is zero when $\mathcal G(x,a)$ is empty.
For $g\in\mathcal G(x,a)$, write $\mathcal C_g=\mathcal C(r_g)$ and $P_g=P_\theta(\mathcal C_g\mid x')$. Selecting a single valid completion $u^\star\in\mathcal C_g$ yields
\begin{equation}
-M_g\log p_\theta(u^\star\mid x')
=
-M_g\log P_g
-M_g\log\frac{p_\theta(u^\star\mid x')}{P_g}.
\label{eq:single-set-decomposition-main}
\end{equation}
The first term encourages completion-set probability; the second favors the selected token within the set. ESCD's child loss retains only the first term per group, supervising valid completions through their aggregate probability without prescribing individual token probabilities. The local gradient analysis examines how this target choice affects agreement with a byte-event reference gradient.

Root and child losses are assigned to their respective prediction positions, with zero contributions where inapplicable, and summed before applying the training mask and reduction. Writing their contributions under this common reduction as $\mathcal L_{\mathrm{root}}$ and $\mathcal L_{\mathrm{child}}$, we obtain
\begin{equation}
\mathcal L_{\mathrm{ESCD}}
=
\mathcal L_{\mathrm{root}}
+
\mathcal L_{\mathrm{child}}.
\label{eq:escd_objective}
\end{equation}
The terms are not separately averaged over eligible positions. Child supervision reuses student logits at visited states without additional rollouts. Algorithm~\ref{alg:framework-escd} summarizes training.
\section{Experiments}
\label{sec:experiments}

\subsection{Experiment Settings}

We use \textbf{Qwen3.5-2B} as the student with two teachers: \textbf{Qwen3-32B}, which differs in tokenizer within the Qwen family, and \textbf{GLM-Z1-9B}, which differs in both model family and tokenizer. Unless otherwise stated, our controlled algorithmic and cross-tokenizer analyses use these pairs. We evaluate mathematical reasoning on AIME 2024--2026 and HMMT 2026~\citep{aime}, and code generation on LiveCodeBench~\citep{livecodebenchmark} and TACO~\citep{TACO}. For large-scale validation, we evaluate \textbf{Qwen3.5-397B-A17B} $\rightarrow$ \textbf{Qwen3-30B-A3B-Thinking} and \textbf{Kimi-K2.7-Code} $\rightarrow$ \textbf{Qwen3.6-35B-A3B} on FrontierScience Olympiad~\citep{frontierScience} and our physics benchmark PHYRD-40. In result tables, \textcolor{red}{\ensuremath{\dagger}} marks teacher scores as references for student performance. Benchmark details and evaluation protocols are in Appendix~\ref{sec:benchmarks-evaluation}.

\begin{table*}[t]
\centering

\caption{
Main results of cross-tokenizer on-policy distillation.
The student is Qwen3.5-2B, with Qwen3-32B and GLM-Z1-9B as teachers.
We report avg@$k$ and pass@$k$ (\%) for each benchmark.
}
\label{tab:main_results_two_teachers}
\vspace{-1.5mm}
\small
\setlength{\tabcolsep}{3.4pt}
\renewcommand{\arraystretch}{1.12}

\begin{adjustbox}{max width=\textwidth}
\begin{tabular}{@{}l*{12}{c}@{}}
\toprule
& \multicolumn{8}{c}{\textbf{Mathematics} ($k=8$)}
& \multicolumn{4}{c}{\textbf{Code} ($k=2$)} \\
\cmidrule(lr){2-9}
\cmidrule(lr){10-13}

\textbf{Method}
& \multicolumn{2}{c}{\textbf{AIME24}}
& \multicolumn{2}{c}{\textbf{AIME25}}
& \multicolumn{2}{c}{\textbf{AIME26}}
& \multicolumn{2}{c}{\textbf{HMMT}-26}
& \multicolumn{2}{c}{\textbf{LiveCodeBench}}
& \multicolumn{2}{c}{\textbf{TACO}} \\
\cmidrule(lr){2-3}
\cmidrule(lr){4-5}
\cmidrule(lr){6-7}
\cmidrule(lr){8-9}
\cmidrule(lr){10-11}
\cmidrule(lr){12-13}

& \textbf{avg@8} & \textbf{pass@8}
& \textbf{avg@8} & \textbf{pass@8}
& \textbf{avg@8} & \textbf{pass@8}
& \textbf{avg@8} & \textbf{pass@8}
& \textbf{avg@2} & \textbf{pass@2}
& \textbf{avg@2} & \textbf{pass@2} \\
\midrule

\rowcolor{gray!10}
\textbf{Qwen3.5-2B (base)}
& 10.4 & 30.0
& 10.0 & 20.0
& 5.0 & 20.0
& 4.2 & 15.2
& 11.5 & 13.7
& 5.3 & 7.1 \\

\midrule

\textbf{Qwen3-32B}{\textcolor{red}{\ensuremath{\dagger}}}
& 79.6 & 90.0
& 69.2 & 83.3
& 72.1 & 90.0
& 27.7 & 48.5
& 59.1 & 65.9
& 53.2 & 56.9 \\

\quad GOLD
& 23.3 & 46.7
& 24.2 & 53.3
& 22.9 & 56.7
& 18.9 & 33.3
& 14.0 & 14.8
& 6.9 & 10.2 \\

\quad X-Token
& 28.8 & 46.7
& 26.7 & 50.0
& 24.2 & 46.7
& 14.0 & 33.3
& 12.6 & \snd{18.1}
& 8.0 & 12.4 \\

\quad SimCT
& 35.0 & \snd{70.0}
& \snd{34.2} & \snd{63.3}
& 34.2 & \snd{63.3}
& 20.8 & \best{42.4}
& 14.8 & 14.8
& 5.5 & 9.9 \\

\quad BPM
& \snd{38.8} & 66.7
& 32.1 & 60.0
& \snd{35.0} & 56.7
& \snd{21.2} & \snd{36.4}
& \snd{15.4} & 15.9
& \snd{11.5} & \snd{18.7} \\

\quad\textbf{Ours}
& \best{46.7} & \best{80.0}
& \best{37.9} & \best{66.7}
& \best{43.8} & \best{66.7}
& \best{27.7} & \best{42.4}
& \best{31.0} & \best{42.9}
& \best{30.0} & \best{47.3} \\

\quad\textcolor{gray}{$\Delta$ vs. best baseline}
& \textcolor{gray}{+7.9}  & \textcolor{gray}{+10.0}
& \textcolor{gray}{+3.7}  & \textcolor{gray}{+3.4}
& \textcolor{gray}{+8.8}  & \textcolor{gray}{+3.4}
& \textcolor{gray}{+6.5}  & \textcolor{gray}{0.0}
& \textcolor{gray}{+15.6} & \textcolor{gray}{+24.8}
& \textcolor{gray}{+18.5} & \textcolor{gray}{+28.6} \\

\midrule

\textbf{GLM-Z1-9B}{\textcolor{red}{\ensuremath{\dagger}}}
& 65.0 & 83.3
& 54.6 & 76.7
& 65.4 & 90.0
& 22.7 & 45.5
& 46.4 & 55.5
& 52.1 & 56.9 \\

\quad GOLD
& 30.8 & \snd{63.3}
& 26.3 & \snd{53.3}
& 27.5 & 56.7
& 12.5 & 27.3
& 13.2 & 17.0
& 9.9 & 14.5 \\

\quad X-Token
& \snd{31.3} & 56.7
& \snd{27.5} & 50.0
& 30.4 & 56.7
& 12.5 & 21.2
& 14.6 & 19.8
& 8.8 & 14.1 \\

\quad SimCT
& \snd{31.3} & \snd{63.3}
& \snd{27.5} & \best{56.7}
& 29.6 & \snd{60.0}
& 19.7 & 33.3
& 14.8 & 17.6
& 9.0 & 13.1 \\

\quad BPM
& 30.8 & 56.7
& 25.4 & 46.7
& \snd{32.1} & 53.3
& \snd{20.8} & \snd{36.4}
& \snd{22.0} & \snd{25.8}
& \snd{16.4} & \snd{23.3} \\

\quad\textbf{Ours}
& \best{35.8} & \best{66.7}
& \best{29.6} & \best{56.7}
& \best{35.0} & \best{70.0}
& \best{25.4} & \best{39.4}
& \best{28.3} & \best{35.2}
& \best{22.6} & \best{29.3} \\

\quad\textcolor{gray}{$\Delta$ vs. best baseline}
& \textcolor{gray}{+4.5} & \textcolor{gray}{+3.4}
& \textcolor{gray}{+2.1} & \textcolor{gray}{0.0}
& \textcolor{gray}{+2.9} & \textcolor{gray}{+10.0}
& \textcolor{gray}{+4.6} & \textcolor{gray}{+3.0}
& \textcolor{gray}{+6.3} & \textcolor{gray}{+9.4}
& \textcolor{gray}{+6.2} & \textcolor{gray}{+6.0} \\

\bottomrule
\end{tabular}
\end{adjustbox}

\vspace{-3.5mm}
\end{table*}

\subsection{Distillation Performance}
\label{sec:distillation-performance}

Table~\ref{tab:main_results_two_teachers} presents the main cross-tokenizer distillation results with Qwen3.5-2B as the student under two different teachers. ESCD consistently outperforms the representative baselines across both settings. With Qwen3-32B as the teacher, ESCD delivers clear gains on all mathematics benchmarks, with improvements of up to $+8.8$ and $+10.0$ points, while the advantage becomes substantially larger on code generation, reaching $+15.6/+24.8$ points on LiveCodeBench and $+18.5/+28.6$ points on TACO. The same trend remains with GLM-Z1-9B: ESCD improves over the strongest baselines across nearly all benchmarks, including gains of $+2.9/+10.0$ points on AIME26, $+6.3/+9.4$ points on LiveCodeBench, and $+6.2/+6.0$ points on TACO. These consistent improvements across two teachers of different model families and capacities demonstrate the robustness of ESCD, with \textbf{particularly strong benefits on code generation}.

Table~\ref{tab:cross_tokenizer_results} extends our dense-model evaluation to heterogeneous MoE distillation at Teacher--Student total-parameter ratios of $13.2\times$ (397B Qwen) and $28.6\times$ (1T Kimi). \textbf{ESCD improves over BPM in both settings}, using direct OPD for Qwen and SFT-initialized OPD for Kimi. In the Kimi setting, direct BPM-based OPD exhibits repetitive continuations and premature termination (Appendix~\ref{app:opd-degeneration}), highlighting a stability challenge in this configuration. Starting from the SFT checkpoint, ESCD further improves FrontierScience Olympiad accuracy from 68.0\% to 72.0\% and the PHYRD-40 mean score from 66.2 to 74.3. These results support effective cross-tokenizer transfer under both initialization regimes and \textbf{the complementary roles of SFT and OPD}: SFT provides the initialization, while subsequent OPD with ESCD yields additional performance gains.

\begin{table}[t]
\centering
\caption{Cross-tokenizer OPD on heterogeneous MoE pairs with 397B and 1T teachers. In the Kimi setting, BPM and ESCD use the same SFT-initialized checkpoint due to direct OPD instability.}
\label{tab:cross_tokenizer_results}

\vspace{-1.5mm}
\small
\setlength{\tabcolsep}{7pt}
\renewcommand{\arraystretch}{1.12}

\begin{tabular}{@{}lcc@{}}
\toprule
\textbf{Method}
& \makecell{\textbf{FrontierScience} \\ \textbf{Olympiad-100}}
& \textbf{PHYRD-40} \\ 
\midrule

\textbf{Qwen3.5-397B-A17B}{\textcolor{red}{\ensuremath{\dagger}}}
& 70.0 & 79.8 \\

\rowcolor{gray!10}
Qwen3-30B-A3B-Thinking (base)
& 48.0 & 43.0 \\

\quad BPM
& 48.0 & 46.6 \\

\quad \textbf{Ours}
& \textbf{51.0} & \textbf{49.6} \\

\quad \textcolor{gray}{$\Delta$ vs. best baseline}
& \textcolor{gray}{+3.0}
& \textcolor{gray}{+3.0} \\

\midrule


\textbf{Kimi-K2.7-Code}{\textcolor{red}{\ensuremath{\dagger}}}
& 75.0 & 83.6 \\

\rowcolor{gray!10}
Qwen3.6-35B-A3B (base)
& 61.0 & 59.2 \\

\rowcolor{gray!10}
Qwen3.6-35B-A3B (SFT)
& 68.0 & 66.2 \\

\quad BPM
& 69.0 & 69.2 \\

\quad \textbf{Ours}
& \textbf{72.0} & \textbf{74.3} \\

\quad \textcolor{gray}{$\Delta$ vs. best baseline}
& \textcolor{gray}{+3.0}
& \textcolor{gray}{+5.1} \\

\bottomrule
\end{tabular}
\vspace{-3.5mm}

\end{table}

\vspace{-2.5mm}

\paragraph{Scope and training stability.}
ESCD addresses supervision across tokenizer boundaries, while OPD performance and stability also depend on student capacity, initialization, training data, and rollout quality. Event-completion supervision therefore complements initialization and optimization strategies rather than providing a general remedy for OPD collapse.
\vspace{-1.5mm}
\section{Ablation and Mechanism Analysis}
\label{sec:escd-ablation}
\vspace{-2mm}

We examine two ESCD questions: \textit{how preserving the \textbf{completion set} affects agreement with a byte-event reference gradient}, and \textit{how frequently \textbf{one-step completion} is available along on-policy student trajectories}. The first compares local objectives on fixed student predictions to isolate their immediate gradient effects; the second measures child-supervision availability.

On matched student contexts and teacher events, we compare three local objectives while holding student predictions, teacher masses, and BPM root supervision fixed. \textbf{Root} uses only root supervision, with no additional loss at the visited child. \textbf{Single} adds a child loss for one deterministically selected valid completion: its negative log-probability is weighted by the teacher event mass. \textbf{Event} uses the same mass weight and visited child, but applies the negative log to the summed probability of all valid one-step completions, without prescribing individual token probabilities. Thus, Root versus Event tests the effect of adding completion-set supervision, while Single versus Event tests the effect of preserving the set rather than selecting one member. Figure~\ref{fig:root-single-event-appendix} illustrates these differences.

We evaluate supervision objectives on previously generated student trajectories without additional model training, using \textbf{Cross-Only Update Fidelity (COUF)} to measure agreement with a specified byte-event reference gradient. Holding student predictions, teacher events and their masses, conditioning contexts, and root supervision fixed enables a controlled comparison at a common model state. The reference differentiates an objective based on event probabilities summed over compatible student paths, including alternative first-token branches and multi-token realizations, whereas Event supervises one-step completion at the visited child. For each context $x$, candidate and reference gradients with respect to student logits at the root and the same depth-one states are concatenated in a common coordinate order into $z_x$ and $z_x^\star$. COUF captures differences in both direction and magnitude by aggregating squared gradient errors across contexts and normalizing by total reference-gradient energy; Appendix~\ref{app:conditioning-state} further examines child-state selection:

\begin{equation}
\operatorname{COUF}
=
1-
\frac{
\sum_x\|z_x-z_x^\star\|_2^2
}{
\sum_x\|z_x^\star\|_2^2
}.
\label{eq:ablation-couf}
\end{equation}

A value of $1$ indicates exact reference agreement, $0$ matches the zero-gradient baseline, and negative values indicate greater error than that baseline. This diagnostic evaluates local logit gradients, not downstream performance after independent training.

\begin{table}[t]
\centering
\small
\setlength{\tabcolsep}{5pt}
\caption{Local target comparison on contexts satisfying the COUF diagnostic criteria. We report reference-energy-weighted state-space COUF; $\Delta$ denotes Event minus Single.}
\vspace{-1.5mm}
\label{tab:event-set-ablation}
\begin{tabular}{lcccc}
\toprule
\textbf{Teacher $\rightarrow$ Student}
& \textbf{Root}
& \textbf{Single}
& \textbf{Event}
& $\boldsymbol{\Delta}$ \\
\midrule
Qwen3 $\rightarrow$ Qwen3.5
& 0.7383
& 0.4323
& \textbf{0.8495}
& +0.4172 \\
GLM-Z1 $\rightarrow$ Qwen3.5
& 0.8848
& 0.7493
& \textbf{0.9341}
& +0.1848 \\
\bottomrule
\end{tabular}
\end{table}

Table~\ref{tab:event-set-ablation} shows higher reference agreement for \textbf{Event} than \textbf{Root}, increasing COUF from 0.7383 to 0.8495 for Qwen and from 0.8848 to 0.9341 for GLM. \textbf{Single} falls below Root in both settings. These results support preserving the completion set over selecting one valid token in the examined contexts. Event need not attain a score of $1$: its root gradient, restriction to the visited child, and local completion objective differ from the reference.

The completion set also admits tokens extending beyond the residual boundary, such as \texttt{eads} alongside \texttt{ead} in Figure~\ref{fig:method}(b). In a separate diagnostic on naturally visited strict-prefix events with available child logits, boundary-crossing tokens account for 2.10\% and 0.63\% of completion probability, respectively. Removing them yields gradients with cosine similarities of 0.9832 and 0.9887 to the full-set gradients. These results indicate that boundary-crossing tokens have limited impact on the local gradient direction in the analyzed subsets.

\vspace{-1.5mm}

\subsection{On-Policy Availability of Child Supervision}
\label{sec:ablation-coverage}
\vspace{-1.5mm}

We examine child-supervision availability using the full teacher vocabulary at aligned, non-whitespace positions in 16 frozen student trajectories per pair. After exact-shared and invalid-token filtering, we aggregate candidates into representative events. Among events admitting a strict-prefix student action, \textbf{Observed} measures the teacher mass fraction whose compatible child is actually visited. Conditional on this visited mass, \textbf{1-step} measures the fraction admitting completion by one additional student token; the remainder is reported as \textbf{Deeper}.

\begin{table}[t]
\centering
\small
\setlength{\tabcolsep}{7pt}
\caption{Full-vocabulary child-supervision availability. Observed is relative to branchable teacher mass; 1-step and Deeper characterize representative residuals within visited mass.}
\vspace{-1.5mm}

\label{tab:escd-coverage}
\begin{tabular}{lccc}
\toprule
\textbf{Teacher $\rightarrow$ Student}
& \textbf{Observed}
& \textbf{1-step}
& \textbf{Deeper} \\
\midrule
Qwen3 $\rightarrow$ Qwen3.5
& 81.83\%
& 99.01\%
& 0.99\% \\
GLM-Z1 $\rightarrow$ Qwen3.5
& 83.29\%
& 99.43\%
& 0.57\% \\
\bottomrule
\end{tabular}
\end{table}

Table~\ref{tab:escd-coverage} shows that compatible children are visited for 81.83\% and 83.29\% of branchable teacher mass for Qwen and GLM, respectively. Conditional on these visited child states, representative residual constraints covering 99.01\% and 99.43\% of the visited teacher mass admit at least one student token that completes the residual bytes in one additional step. These results show that after partial event entry, the vast majority of observed residual constraints are covered by ESCD's one-step child supervision, supporting the practical scope of the proposed objective.
\vspace{-1.5mm}

\section{Conclusion}
\label{sec:conclusion}
\vspace{-1.5mm}
We identify the \textbf{event-completion gap} in cross-tokenizer on-policy distillation: a student action can enter a teacher byte event while leaving a residual byte constraint whose completion is not uniquely represented by the student tokenizer. We introduce \textbf{Event-Set Completion Distillation (ESCD)}, which complements root-level alignment with completion-set supervision at naturally visited student child states. ESCD aggregates prefix-related teacher events and supervises the total probability of byte-compatible student completions, without prescribing individual completion probabilities or requiring counterfactual rollouts. Experiments show consistent gains across mathematics, code, and scientific reasoning, including heterogeneous MoE distillation from a 1T teacher to a 35B student. Local analyses support preserving the completion set, while one-step-completable residual constraints cover over 99\% of observed compatible teacher mass. These findings support event entry and event completion as complementary supervision targets for cross-tokenizer knowledge transfer.
\newpage
\subsubsection*{AI Use Statement}
In this work, we used generative AI tools for language polishing, translation assistance for author-written drafts, and limited code assistance (e.g., debugging and refactoring of training and evaluation scripts). We did not use generative AI tools to generate experimental results, fabricate data, or make final scientific claims. Separately from writing assistance, large language models appear in this work as research objects: all teacher and student models are publicly available checkpoints studied in our experiments. In addition, following common practice for open-ended scientific answers, we use DeepSeek-V4 Pro Preview as an automatic grader for FrontierScience Olympiad and PHYRD-40, conditioned on reference answers, reference solutions, and problem-specific rubrics; the grading protocol is described in Appendix~\ref{sec:benchmarks-evaluation}, and the same grader and prompts are applied to all compared models. All AI-assisted text, code, and suggestions were manually reviewed, revised, and verified by the authors. We take full responsibility for the final content of this work, including text, claims, and artifacts produced with the aid of generative AI.

\subsubsection*{Ethics Statement}
This work studies cross-tokenizer on-policy distillation, a technique for transferring capabilities from a teacher language model to a student with a different tokenizer. Such techniques can reduce the cost of deploying capable reasoning models, but distillation may also transfer undesirable behaviors of the teacher, including factual errors, social biases, and unsafe outputs; our evaluation focuses on reasoning accuracy and does not assess safety or bias, so distilled models should undergo separate safety evaluation before deployment. Cross-tokenizer distillation may further be used to replicate capabilities of models whose terms of use restrict distillation; all teacher and student models in this work are publicly released checkpoints used within their license terms, and we encourage practitioners to respect the licenses of the models they distill. Our training and evaluation data consist of publicly available mathematics, code, and science benchmarks together with SciDeriv, which is derived from mathematical and scientific documents and contains prompts only. These data do not involve personally identifiable information, and our work does not involve human-subject experiments. The PHYRD-40 problems were written by invited domain scientists, and only problem statements, reference solutions, and rubrics are used. Model-generated code is executed only in isolated sandboxed environments. Finally, our experiments require substantial GPU resources (Table~\ref{tab:training-settings}); ESCD itself reuses on-policy trajectories and requires no additional rollouts.

\subsubsection*{Reproducibility Statement}
We support reproducibility through detailed descriptions of our method, training and evaluation pipeline, and data in the main paper and appendix. Section~\ref{sec:method} specifies the residual-event formulation, completion-set construction, and training objective, while Appendix~\ref{sec:framework-escd} provides further details, Appendix~\ref{sec:framework-worked-example} gives a worked example, and Algorithm~\ref{alg:framework-escd} summarizes the training procedure. Appendix~\ref{sec:framework-existing-methods} presents the compared baselines under a unified view, and Appendix~\ref{sec:training-framework} describes their shared implementation framework. Chat-template rendering, retokenization, and special-token handling are detailed in Appendix~\ref{sec:template-special-token}. Key OPD hyperparameters, rollout settings, and hardware configurations are summarized in Table~\ref{tab:training-settings}; training data composition and benchmark filtering are described in Appendices~\ref{sec:training-data} and~\ref{sec:benchmarks-evaluation}; and decoding settings are given in Table~\ref{tab:generation-settings}. All teacher and student models are initialized from publicly available checkpoints. We will release our code, including the ESCD and baseline plugins, the SciDeriv prompts, the PHYRD-40 problem set with reference solutions and scoring rubrics, and the evaluation scripts and grading prompts to support reproduction of the reported results.



\clearpage
\bibliography{iclr2027_conference}
\bibliographystyle{iclr2027_conference}
\clearpage
\appendix

\clearpage
\section{Training Framework and Settings}
\label{app:training-framework-settings}

This appendix documents the training and evaluation implementation used throughout our experiments. We organize the details as follows:

\begin{enumerate}
    \item \textbf{Training framework} (\ref{sec:training-framework}): on-policy rollout, Teacher serving, Student optimization, and the unified implementation of cross-tokenizer baselines.

    \item \textbf{Cross-model chat templates and special-token handling} (\ref{sec:template-special-token}): model-native prompt rendering, response retokenization, EOS bridging, truncation, and thinking markers.

    \item \textbf{Training settings} (\ref{sec:training-settings}): optimization hyperparameters, rollout budgets, and distributed-training configuration.

    \item \textbf{Training data} (\ref{sec:training-data}): the mathematics and code prompts used for distillation.

    \item \textbf{Benchmarks and evaluation} (\ref{sec:benchmarks-evaluation}): benchmark construction and task-specific grading protocols.

    \item \textbf{Common generation protocol} (\ref{sec:generation-protocol}): decoding settings, sampling budgets, response-length limits, and truncation rules shared across compared models.
\end{enumerate}

\subsection{Training Framework}
\label{sec:training-framework}

We conduct all on-policy distillation experiments with Slime~\citep{slime}, which integrates Student rollout generation, Teacher scoring, and distributed Student optimization. At each step, the current Student generates responses with SGLang; these responses are retokenized and scored by a frozen Teacher, and only the Student is updated using the resulting cross-tokenizer supervision.

BPM, SimCT, GOLD, X-Token, and ESCD are implemented through a unified plugin interface. Within each Teacher--Student pair, all methods share the training data, rollout settings, and optimization budget, while each generates trajectories with its own Student and obtains Teacher predictions on those trajectories. The plugins implement method-specific alignment and supervision objectives. The Teacher is served with SGLang, while the Student is optimized with Megatron-LM~\citep{megatron-lm} through Slime's distributed actor backend. This setup provides a common infrastructure for rollout, scheduling, optimization, checkpointing, and evaluation while preserving on-policy training for each method.

\subsection{Cross-Model Chat Templates and Special-Token Handling}
\label{sec:template-special-token}

Cross-family distillation introduces implementation details absent when Teacher and Student share a tokenizer and chat template. Directly copying token IDs or serialized prompts across models can misalign role boundaries, duplicate termination symbols, or misrepresent truncated responses as completed. We therefore separate two operations: (1) each model renders the structured conversation with its native chat template, and (2) model-specific special tokens are handled through explicit semantic rules rather than token-ID equality.

For the three model families used in our experiments, the native chat templates differ substantially in their role and control tokens. Figure~\ref{fig:cross-model-chat-templates} shows their template structures. In all cases, we retain the original structured conversation and let each model render it using its own native chat template with thinking enabled. We never translate model-specific template tokens between tokenizers.

\begin{figure*}[hbt]
\centering

\begin{minipage}[t]{0.315\textwidth}
\vspace{0pt}
\begin{tcolorbox}[
    enhanced,
    colback=boxgray,
    colframe=linegray,
    colbacktitle=red!10,
    coltitle=black,
    fonttitle=\bfseries\small,
    fontupper=\ttfamily\scriptsize,
    title={Qwen family},
    boxrule=0.5pt,
    arc=1.5mm,
    outer arc=1.5mm,
    left=2.8mm,
    right=2.8mm,
    top=2.5mm,
    bottom=2.5mm,
    toptitle=2.3mm,
    bottomtitle=2.3mm,
    boxsep=0pt,
]

\textcolor{qwenblue}{<|im\_start|>system}\\
\textcolor{textgray}{\normalfont\itshape\{system message\}}\\
\textcolor{qwenblue}{<|im\_end|>}

\vspace{1.5mm}

\textcolor{qwenblue}{<|im\_start|>user}\\
\textcolor{textgray}{\normalfont\itshape\{user message\}}\\
\textcolor{qwenblue}{<|im\_end|>}

\vspace{1.5mm}

\textcolor{qwenblue}{<|im\_start|>assistant}\\
\textcolor{qwenblue}{<think>}\textsuperscript{\normalfont *}\\
\textcolor{textgray}{\normalfont\itshape\{generated text\}}

\end{tcolorbox}
\end{minipage}
\hfill
\begin{minipage}[t]{0.315\textwidth}
\vspace{0pt}
\begin{tcolorbox}[
    enhanced,
    colback=boxgray,
    colframe=linegray,
    colbacktitle=glmviolet!12,
    coltitle=black,
    fonttitle=\bfseries\small,
    fontupper=\ttfamily\scriptsize,
    title={GLM-Z1-9B},
    boxrule=0.5pt,
    arc=1.5mm,
    outer arc=1.5mm,
    left=2.8mm,
    right=2.8mm,
    top=2.5mm,
    bottom=2.5mm,
    toptitle=2.3mm,
    bottomtitle=2.3mm,
    boxsep=0pt,
]

\textcolor{glmviolet}{[gMASK]<sop>}

\vspace{1.5mm}

\textcolor{glmviolet}{<|system|>}\\
\\
\textcolor{textgray}{\normalfont\itshape\{system message\}}

\vspace{1.5mm}

\textcolor{glmviolet}{<|user|>}\\
\\
\textcolor{textgray}{\normalfont\itshape\{user message\}}

\vspace{1.5mm}

\textcolor{glmviolet}{<|assistant|>}\\
\textcolor{textgray}{\normalfont\itshape\{generated text\}}

\end{tcolorbox}
\end{minipage}
\hfill
\begin{minipage}[t]{0.315\textwidth}
\vspace{0pt}
\begin{tcolorbox}[
    enhanced,
    colback=boxgray,
    colframe=linegray,
    colbacktitle=kimigreen!12,
    coltitle=black,
    fonttitle=\bfseries\small,
    fontupper=\ttfamily\scriptsize,
    title={Kimi-K2.7-Code},
    boxrule=0.5pt,
    arc=1.5mm,
    outer arc=1.5mm,
    left=2.8mm,
    right=2.8mm,
    top=2.5mm,
    bottom=2.5mm,
    toptitle=2.3mm,
    bottomtitle=2.3mm,
    boxsep=0pt,
]

\textcolor{kimigreen}{<|im\_system|>system}\\
\textcolor{kimigreen}{<|im\_middle|>}%
\textcolor{textgray}{\normalfont\itshape\{system message\}}\\
\textcolor{kimigreen}{<|im\_end|>}

\vspace{1.5mm}

\textcolor{kimigreen}{<|im\_user|>user}\\
\textcolor{kimigreen}{<|im\_middle|>}%
\textcolor{textgray}{\normalfont\itshape\{user message\}}\\
\textcolor{kimigreen}{<|im\_end|>}

\vspace{1.5mm}

\textcolor{kimigreen}{<|im\_assistant|>assistant}\\
\textcolor{kimigreen}{<|im\_middle|><think>}\\
\textcolor{textgray}{\normalfont\itshape\{generated text\}}

\end{tcolorbox}
\end{minipage}

\vspace{1mm}
{\scriptsize
\textsuperscript{*}The Qwen3-32B template omits this opening
\texttt{<think>} in thinking mode.
\par}

\caption{
\textbf{Native chat-template prefixes across model families.}
We illustrate a text-only conversation without tool calls, using each checkpoint's native generation prefix in its thinking configuration. Colored text denotes template-supplied prefixes; \textit{generated text} begins after the displayed prefix. GLM-Z1-9B and Qwen3-32B do not prefill an opening \texttt{<think>}, whereas the other illustrated thinking configurations do. Model-specific role and control tokens remain local to their respective tokenizers; Student-generated response text is retokenized for Teacher scoring. Layout line breaks are schematic; exact whitespace follows each checkpoint's template.
}
\label{fig:cross-model-chat-templates}
\end{figure*}

\subsubsection{Special-Token Handling}
\label{sec:special-token-handling}

Different model families use different tokens to mark the end of a response. We follow each model's native generation semantics rather than assuming EOS token IDs or surface forms are shared across tokenizers. In particular, we recognize tokenizer-declared EOS tokens together with stop tokens declared by the model's generation configuration, while other special tokens are not treated as terminal by default. Table~\ref{tab:model-special-tokens} summarizes the model-specific termination rules used in our experiments.
\begin{table}[hbt]
\centering
\caption{
Model-specific termination handling. The bridge stop is the token used to represent a normally completed response after cross-tokenizer retokenization.
}
\vspace{-3mm}
\label{tab:model-special-tokens}
\small
\setlength{\tabcolsep}{5pt}
\renewcommand{\arraystretch}{1.15}

\begin{tabular}{@{}lll@{}}
\toprule
\textbf{Model}
& \textbf{Recognized termination}
& \textbf{Preferred bridge stop} \\
\midrule
Qwen3 / Qwen3.5
& model-declared EOS / stop tokens
& \texttt{<|im\_end|>} \\

GLM-Z1
& \texttt{<|endoftext|>} and configured stops
& \texttt{<|endoftext|>} \\

Kimi-K2.7-Code
& \texttt{[EOS]}, \texttt{<|im\_end|>}
& \texttt{<|im\_end|>} \\
\bottomrule
\end{tabular}
\end{table}

\paragraph{Termination and truncation.}
When a Student rollout ends with one of its recognized termination tokens, we remove that final token before decoding and retokenizing the response content for the Teacher. After retokenization, a Teacher-native bridge stop is appended to preserve the fact that the original response terminated normally. The bridge token is model specific: for example, GLM-Z1 uses \texttt{<|endoftext|>}, whereas Kimi-K2.7-Code uses \texttt{<|im\_end|>} even though its tokenizer also defines \texttt{[EOS]}. In contrast, if generation reaches the maximum response length without producing a termination token, the response is treated as truncated and no EOS or bridge stop is added.

During cross-tokenizer supervision, these model-specific stop tokens are treated as the same semantic terminal event rather than as ordinary lexical content. Thus, termination can be aligned across models without requiring their EOS token IDs or textual forms to match. GLM-Z1 additionally declares role or interaction boundaries such as \texttt{<|user|>} and \texttt{<|observation|>} as generation stops; we respect these declarations when determining whether a rollout has terminated.

\paragraph{Thinking markers.}
We follow each checkpoint's native thinking-mode generation prefix (Figure~\ref{fig:cross-model-chat-templates}). Template prefilling of \texttt{<think>} is checkpoint-specific: Qwen3-32B and GLM-Z1-9B do not prefill it, whereas the other illustrated configurations do. We do not impose a common opening marker during cross-model conversion. Any generated \texttt{<think>} or \texttt{</think>} is retained and retokenized with the surrounding text for Teacher scoring. Neither marker is treated as a termination event unless explicitly declared by the model configuration. If generation is truncated before \texttt{</think>}, we do not insert a closing marker. Prompt-side role and control markers remain model-specific, while generated reasoning markers are transferred as response text rather than copied as tokenizer-local IDs.

\subsection{Training Settings} 
\label{sec:training-settings} 
 
Table~\ref{tab:training-settings} summarizes the configurations for small-scale and large-scale OPD experiments. The small-scale experiments use Qwen3.5-2B as the Student with Qwen3-32B or GLM-Z1-9B as the Teacher. The large-scale experiments cover Qwen3.5-397B-A17B to Qwen3-30B-A3B-Thinking and Kimi-K2.7-Code to Qwen3.6-35B-A3B distillation. For the Kimi-K2.7-Code to Qwen3.6-35B-A3B setting, both BPM and ESCD are applied on the same SFT-initialized Student checkpoint to ensure a fair comparison. Within each Teacher--Student pair, compared methods share the training data, rollout budget, optimizer settings, and distributed training configuration, while differing in their cross-tokenizer objectives. Training datasets are described in Section~\ref{sec:training-data}. 
\begin{table*}[t]
\centering
\caption{Training configurations for OPD experiments. GPU totals include Teacher serving, Student optimization, and Student rollout, excluding separately scheduled evaluation jobs.}

\label{tab:training-settings}
\small
\setlength{\tabcolsep}{5pt}
\renewcommand{\arraystretch}{1.12}
\begin{tabular}{@{}l|c|cc@{}}
\toprule
\textbf{Setting} & \textbf{Small-scale} & \multicolumn{2}{c}{\textbf{Large-scale}} \\
\midrule
Teacher & Qwen3-32B / GLM-Z1-9B & Qwen3.5-397B-A17B & Kimi-K2.7-Code \\
Student & Qwen3.5-2B & Qwen3-30B-A3B-Thinking & Qwen3.6-35B-A3B \\
Training framework & Slime & Slime & Slime \\
Student optimization & Megatron-LM & Megatron-LM & Megatron-LM \\
Rollout and Teacher serving & SGLang & SGLang & SGLang \\
Training data & DAPO-Math + TACO & SciDeriv & SciDeriv \\
Number of training prompts & $20{,}000$ & $17{,}577$ & $17{,}577$ \\
Global rollout batch size & $16$ & $16$ & $16$ \\
Global training batch size & $16$ & $16$ & $16$ \\
Responses per prompt & $1$ & $1$ & $1$ \\
Micro-batch size & $1$ & $1$ & $1$ \\
Maximum prompt length & $5{,}120$ tokens & $4{,}096$ tokens& $4{,}096$ tokens\\
Maximum response length & $27{,}648$ tokens& $27{,}648$ tokens& $80{,}000$ tokens\\
Maximum sequence length & $33{,}280$ tokens& $33{,}280$ tokens& $84{,}112$ tokens\\
Rollout temperature & $1.0$ & $1.0$ & $1.0$ \\
Rollout top-$p$ & $0.95$ & $0.95$ & $0.95$ \\
Rollout top-$k$ & $20$ & $20$ & $20$ \\
Learning rate & $5\times10^{-7}$ & $1\times10^{-6}$ & $1\times10^{-6}$ \\
Weight decay & $0.1$ & $0.1$ & $0.1$ \\
Tensor parallel size & $2$ & $2$ & $2$ \\
Pipeline parallel size & $2$ & $4$ & $2$ \\
Context parallel size & $8$ & $8$ & $8$ \\
Checkpoint interval & $50$ steps & $10$ steps & $10$ steps \\
Evaluation interval & $50$ steps & $10$ steps & $10$ steps \\
\midrule
Teacher GPUs & $16$ & $32$ & $32$ \\
Student training GPUs & $32$ & $64$ & $64$ \\
Student rollout GPUs & $16$ & $32$ & $32$ \\
Total GPUs & $64$ & $128$ & $128$ \\
\bottomrule
\end{tabular}
\end{table*}

\subsection{Training Data}
\label{sec:training-data}

We use two training corpora depending on model scale. For small-scale distillation, we use a balanced mathematics--code mixture of 20,000 prompts: 10,000 DAPO mathematics problems with exact integer answers~\citep{dapo} and 10,000 TACO programming problems with recovered executable test suites~\citep{TACO}. Table~\ref{training_data} summarizes the composition.

For large-scale distillation, we construct \textbf{SciDeriv}, a scientific derivation reasoning corpus derived from mathematical and scientific documents. A document-to-prompt pipeline extracts formulas and intermediate derivation relations from source PDFs and converts them into structured reasoning problems. After filtering, SciDeriv contains 17577 prompts covering two complementary tasks: conclusion assessment and conclusion reconstruction (Table~\ref{training_data_formula}).

In the 10000 conclusion-assessment prompts, a candidate conclusion is provided, and the model must determine whether it is entailed by the supplied relations, merely compatible with them, or inconsistent with them. In the 7577 conclusion-reconstruction prompts, the endpoint is withheld, and the model must reconstruct the strongest conclusion justified by the available relations.

\begin{table}[hbt]
\centering

\begin{minipage}[t]{0.36\textwidth}
\centering
\small
\captionof{table}{Training data for small-scale distillation.}

\label{training_data}
\begin{tabular}{@{}lcc@{}}
\toprule
Task & Mathematics & Code \\
\midrule
Source  & DAPO-Math & TACO \\
Prompts & 10,000    & 10,000 \\
\bottomrule
\end{tabular}
\end{minipage}
\hfill
\begin{minipage}[t]{0.62\textwidth}
\centering
\small
\captionof{table}{Composition of SciDeriv for large-scale distillation.}

\label{training_data_formula}
\begin{tabular}{@{}lc@{}}
\toprule
Task type & Prompts \\
\midrule
Conclusion assessment     & 10,000 \\
Conclusion reconstruction & 7,577 \\
\midrule
Total                     & 17,577 \\
\bottomrule
\end{tabular}
\end{minipage}

\end{table}

Both tasks require reasoning over intermediate transformations, including identifying relevant assumptions, ambiguities, normalization choices, and convention-dependent steps. They also require recognizing when the supplied information is insufficient to determine a unique conclusion. SciDeriv contains prompts only, without stored reference completions or scalar reward labels.

\tcbset{
trainingcase/.style={
    enhanced,
    colback=boxgray,
    colframe=linegray,
    colbacktitle=qwenblue!12,
    coltitle=black,
    fonttitle=\bfseries\small,
    fontupper=\ttfamily\scriptsize,
    boxrule=0.5pt,
    arc=1.5mm,
    outer arc=1.5mm,
    left=2.8mm,
    right=2.8mm,
    top=2.5mm,
    bottom=2.5mm,
    toptitle=2.3mm,
    bottomtitle=2.3mm,
    boxsep=0pt,
}
}

\begin{tcolorbox}[
    trainingcase,
    title={Case 1: Endpoint exposed and verifiable}
]

\textcolor{qwenblue}{<|im\_start|>system}\\
\textcolor{textgray}{\normalfont\itshape
You are a rigorous mathematical reasoning assistant. Analyze the supplied derivation, determine whether the proposed endpoint follows from the given relations, and explicitly identify any required assumptions or conventions.}\\
\textcolor{qwenblue}{<|im\_end|>}

\vspace{1.5mm}

\textcolor{qwenblue}{<|im\_start|>user}\\
\textcolor{textgray}{\normalfont
Consider the phase convention}
\[
S(\theta)=\epsilon e^{i\delta(\theta)},
\qquad
\epsilon\in\{+1,-1\},
\]
\textcolor{textgray}{\normalfont together with}
\[
\rho(i\pi-\theta)=\rho(i\pi+\theta),
\]
\[
\rho(\theta)-\rho(-\theta)
=
i\delta(\theta)
+
\frac{i\pi(1-\epsilon)}{2}
\operatorname{sign}(\theta).
\]

\textcolor{textgray}{\normalfont
Define}
\[
\log F_{\min}(\theta)=\rho(\theta).
\]

\textcolor{textgray}{\normalfont
Determine whether the following endpoint relations are implied by the displayed identities:}
\[
F_{\min}(\theta)
=
F_{\min}(2\pi i-\theta),
\qquad
F_{\min}(\theta)
=
S(\theta)F_{\min}(-\theta).
\]

\textcolor{textgray}{\normalfont
Explain the derivation carefully and state any branch, normalization, or sign conventions needed for the conclusion.}\\
\textcolor{qwenblue}{<|im\_end|>}

\vspace{1.5mm}

\textcolor{qwenblue}{<|im\_start|>assistant}\\
\textcolor{qwenblue}{<think>}\\
\textcolor{textgray}{\normalfont\itshape
\{model-generated reasoning begins here; no reference completion is stored\}}

\end{tcolorbox}

\begin{tcolorbox}[
    trainingcase,
    title={Case 2: Endpoint withheld and reconstructed}
]

\textcolor{qwenblue}{<|im\_start|>system}\\
\textcolor{textgray}{\normalfont\itshape
You are a rigorous mathematical reasoning assistant. Reconstruct the strongest conclusion justified by the supplied derivation, explain which intermediate relations support each part of the result, and identify any remaining ambiguity or missing condition.}\\
\textcolor{qwenblue}{<|im\_end|>}

\vspace{1.5mm}

\textcolor{qwenblue}{<|im\_start|>user}\\
\textcolor{textgray}{\normalfont
Consider}
\[
S(\theta)=\epsilon e^{i\delta(\theta)},
\qquad
\epsilon\in\{+1,-1\},
\]
\textcolor{textgray}{\normalfont and the identities}
\[
\rho(i\pi-\theta)=\rho(i\pi+\theta),
\]
\[
\rho(\theta)-\rho(-\theta)
=
i\delta(\theta)
+
\frac{i\pi(1-\epsilon)}{2}
\operatorname{sign}(\theta).
\]

\textcolor{textgray}{\normalfont
Let}
\[
\log F_{\min}(\theta)=\rho(\theta).
\]

\textcolor{textgray}{\normalfont
Using only these relations, reconstruct the functional equations satisfied by \(F_{\min}\). For each equation, identify the intermediate relation from which it follows. Then determine whether the information above uniquely specifies \(F_{\min}\), and state the minimal additional condition required if it does not.}\\
\textcolor{qwenblue}{<|im\_end|>}

\vspace{1.5mm}

\textcolor{qwenblue}{<|im\_start|>assistant}\\
\textcolor{qwenblue}{<think>}\\
\textcolor{textgray}{\normalfont\itshape
\{model-generated reasoning begins here; no reference completion is stored\}}

\end{tcolorbox}

\paragraph{Representative case studies.}

The assistant field marks only the generation boundary used during training and does not contain a stored reference response. Cases~1 and~3 illustrate the endpoint-exposed setting with different logical statuses of the proposed conclusion, whereas Case~2 illustrates endpoint reconstruction when the conclusion is withheld.

Taken together, the examples highlight the central objective of our formula-reasoning corpus: supervising the \emph{logical relationship} between a derivation and its conclusion rather than only the surface correctness of the final formula. Endpoint-exposed prompts test whether a conclusion is entailed, compatible, or inconsistent with the premises, while endpoint-withheld prompts test reconstruction of the strongest justified conclusion. Both settings require tracking intermediate transformations, implicit assumptions, solution ambiguities, and convention-dependent conditions.

\begin{tcolorbox}[
    trainingcase,
    title={Case 3: Consistent but not entailed}
]

\textcolor{qwenblue}{<|im\_start|>system}\\
\textcolor{textgray}{\normalfont\itshape
You are a rigorous mathematical reasoning assistant. Distinguish carefully between conclusions that are logically entailed by the derivation, conclusions that are merely compatible with it, and conclusions that contradict it. Explicitly identify missing assumptions.}\\
\textcolor{qwenblue}{<|im\_end|>}

\vspace{1.5mm}

\textcolor{qwenblue}{<|im\_start|>user}\\
\textcolor{textgray}{\normalfont
Consider the differential equation}
\[
\frac{dy}{dx}=2xy.
\]

\textcolor{textgray}{\normalfont
For a nonzero solution, separation of variables gives}
\[
\frac{1}{y}\,dy=2x\,dx,
\]
\textcolor{textgray}{\normalfont and hence}
\[
\log |y|=x^2+C.
\]

\textcolor{textgray}{\normalfont
A proposed endpoint is}
\[
y(x)=e^{x^2}.
\]

\textcolor{textgray}{\normalfont
Determine whether this endpoint is entailed by the derivation, merely consistent with it, or inconsistent with it. Give the strongest solution family supported by the available information, identify any additional condition needed to recover the proposed endpoint, and discuss whether the derivation omits any special solution.}\\
\textcolor{qwenblue}{<|im\_end|>}

\vspace{1.5mm}

\textcolor{qwenblue}{<|im\_start|>assistant}\\
\textcolor{qwenblue}{<think>}\\
\textcolor{textgray}{\normalfont\itshape
\{model-generated reasoning begins here; no reference completion is stored\}}

\end{tcolorbox}

\subsection{Benchmarks and Evaluation Protocol}
\label{sec:benchmarks-evaluation}

We evaluate mathematical reasoning, code generation, and scientific reasoning on fixed benchmark sets. The problem sets and evaluation protocols are held constant across all models being compared.

\paragraph{AIME 2024, AIME 2025, and AIME 2026.}\citep{aime}
Each AIME split contains 30 olympiad-style problems with reference answers given as integers in $[0,999]$. We sample eight independent responses per problem. A response is considered correct if the integer extracted from its final answer matches the reference answer.

\paragraph{HMMT February 2026.}\citep{aime}
HMMT contains 33 contest problems. We sample eight independent responses per problem. Unlike AIME, reference answers are not restricted to integers; the answer checker therefore accepts numerically or symbolically equivalent forms, including fractions, radicals, and other closed-form expressions.

\paragraph{LiveCodeBench.}\citep{livecodebenchmark}
We use a fixed January--April 2025 slice containing 182 programming problems and sample two responses per problem. From each response, we extract the last syntactically valid Python program enclosed in a fenced code block and execute it against the official test cases in an isolated filesystem jail. A sample is considered correct only if it passes all required tests within the benchmark resource limits.

\paragraph{TACO.}\citep{TACO}
We use the test split at revision \texttt{d593ed0a}. Starting from 400 easy- and medium-difficulty candidates, we remove 75 special-judge problems, 21 problems with image dependencies or empty test suites, and 21 problems overlapping with the training pool, leaving 283 problems (149 easy and 134 medium). We execute the model-generated Python programs in an isolated environment. Each problem uses at most 40 paired input--output tests, and a solution is considered correct only if its outputs exactly match the expected outputs on all applicable tests.

\paragraph{FrontierScience Olympiad-100.} FrontierScience~\citep{frontierScience} evaluates scientific reasoning across physics, chemistry, and biology. We use its 100-problem Olympiad gold set, consisting of 50 physics, 40 chemistry, and 10 biology problems, and exclude the Research track. We sample one response per problem and use DeepSeek-V4 Pro Preview with the benchmark's reference-answer grading prompt to determine whether each response matches or is equivalent to the reference answer. Each response receives a binary correctness label.

\paragraph{PHYRD-40.} PHYRD-40 comprises 40 physics problems developed by scientists invited by our team, covering quantum field theory and symmetries, gravity and holography, string theory and brane dynamics, celestial scattering amplitudes, cosmological large-scale structure, and tensor-network methods. These problems require sustained analytical reasoning and detailed derivations. We sample one response per problem and use DeepSeek-V4 Pro Preview to evaluate it against the reference solution and problem-specific rubric. Each response receives a score from 0 to 100, and we report the arithmetic mean over 40 problems. We will publicly release the PHYRD-40 problem set, reference solutions, scoring rubrics, and evaluation pipeline.

\subsection{Common Generation Protocol}
\label{sec:generation-protocol}

Within each experimental setting, we apply the same decoding settings and sampling budget to the base Student, the Teacher, and all distilled models. Each problem is formatted with a benchmark-specific instruction and rendered using the evaluated model's native chat template with thinking enabled. We use temperature $0.6$ for mathematics and code, and $1.0$ for scientific reasoning; top-$p=0.95$ and top-$k=20$ are shared across all benchmarks.

Table~\ref{tab:generation-settings} summarizes the generation settings. For AIME, HMMT, LiveCodeBench, and TACO, we use a context limit of $32{,}768$ tokens and an output limit of $27{,}648$ tokens, sampling eight responses per mathematics problem and two per code problem. For FrontierScience Olympiad and PHYRD-40, we use a context limit of $262{,}144$ tokens and an output limit of $256{,}000$ tokens, sampling one response per problem. Token limits are measured using each model's native tokenizer, and the available output budget is also constrained by the remaining context capacity. Within each benchmark, all compared models use the same problem order and grading procedure.

\begin{table}[hbt]
\centering
\caption{Generation settings for the controlled and large-scale experiments. Context and output limits are measured in tokens; the context limit includes both the input prompt and generated response.}
\vspace{-3mm}
\label{tab:generation-settings}
\small
\setlength{\tabcolsep}{4.5pt}
\renewcommand{\arraystretch}{1.12}
\begin{tabular}{@{}lcccccc@{}}
\toprule
\textbf{Evaluation setting}
& \textbf{Context}
& \textbf{Max output}
& \textbf{Temp}.
& \textbf{Top-$p$}
& \textbf{Top-$k$}
& \textbf{Samples/problem} \\
\midrule
AIME24/25/26
& $32{,}768$ & $27{,}648$ & $0.6$ & $0.95$ & $20$ & $8$ \\
HMMT-26
& $32{,}768$ & $27{,}648$ & $0.6$ & $0.95$ & $20$ & $8$ \\
LiveCodeBench
& $32{,}768$ & $27{,}648$ & $0.6$ & $0.95$ & $20$ & $2$ \\
TACO
& $32{,}768$ & $27{,}648$ & $0.6$ & $0.95$ & $20$ & $2$ \\
FrontierScience Olympiad
& $262{,}144$ & $256{,}000$ & $1.0$ & $0.95$ & $20$ & $1$ \\
PHYRD-40
& $262{,}144$ & $256{,}000$ & $1.0$ & $0.95$ & $20$ & $1$ \\
\bottomrule
\end{tabular}
\end{table}

\paragraph{Mathematical reasoning.}
The prompt requests a step-by-step solution ending with an \texttt{Answer:} line. The grader extracts the last such answer or, if absent, the last \verb|\boxed{}| expression, checking numeric and symbolic equivalence.

\begin{tcolorbox}[
    trainingcase,
    title={Mathematical reasoning prompt (used for training and evaluation)}
]

\textcolor{qwenblue}{<|im\_start|>user}\\
{\color{textgray}\normalfont
Solve the following math problem step by step. The last line of your response should be of the form \texttt{Answer: \$Answer} (without quotes), where \texttt{\$Answer} is the answer to the problem.

\medskip

\{problem\}

\medskip

Remember to put your answer on its own line after \texttt{Answer:}.
}
\\
\textcolor{qwenblue}{<|im\_end|>}

\vspace{1.5mm}

\textcolor{qwenblue}{<|im\_start|>assistant}\\
\textcolor{qwenblue}{<think>}\\
{\color{textgray}\normalfont\itshape
\{model-generated reasoning begins here; no reference completion is stored\}
}

\end{tcolorbox}

\paragraph{Code generation.}
The prompt requests a complete Python program in a fenced code block, following the provided function signature or otherwise using standard input and output.
\begin{tcolorbox}[
    trainingcase,
    title={Code generation prompt (used for training and evaluation)}
]

\textcolor{qwenblue}{<|im\_start|>user}\\
{\color{textgray}\normalfont
You are given a competitive programming problem. Think step by step, then provide a single complete Python program in a \texttt{python} code block. If starter code or a function signature is given, implement it; otherwise, read from standard input and write to standard output.

\medskip

\{problem\}
}
\\
\textcolor{qwenblue}{<|im\_end|>}

\vspace{1.5mm}

\textcolor{qwenblue}{<|im\_start|>assistant}\\
\textcolor{qwenblue}{<think>}\\
{\color{textgray}\normalfont\itshape
\{model-generated reasoning begins here; no reference completion is stored\}
}

\end{tcolorbox}

\paragraph{FrontierScience Olympiad.}
The system message requests a rigorous, self-contained scientific solution with supporting reasoning. The user message supplies the subject and problem statement.

\begin{tcolorbox}[
    trainingcase,
    title={FrontierScience Olympiad prompt (used for evaluation)}
]

\textcolor{qwenblue}{<|im\_start|>system}\\
{\color{textgray}\normalfont\itshape
You are an expert scientist. Solve the problem rigorously and return a complete, self-contained answer. Show the reasoning needed to support the final answer.
}
\\
\textcolor{qwenblue}{<|im\_end|>}

\vspace{1.5mm}

\textcolor{qwenblue}{<|im\_start|>user}\\
{\color{textgray}\normalfont
\# Frontier Science Problem

\medskip

Subject: \{subject\}

\medskip

Problem:

\medskip

\{problem\}

\medskip

Provide your complete solution and final answer directly in this response.
}
\\
\textcolor{qwenblue}{<|im\_end|>}

\vspace{1.5mm}

\textcolor{qwenblue}{<|im\_start|>assistant}\\
\textcolor{qwenblue}{<think>}\\
{\color{textgray}\normalfont\itshape
\{model-generated reasoning begins here; no reference completion is stored\}
}

\end{tcolorbox}

\paragraph{PHYRD-40.}
The system message requests a complete derivation with explicit conventions and consistency checks, without external tools or sources. The user message supplies the problem identifier and statement.

\begin{tcolorbox}[
    trainingcase,
    title={PHYRD-40 prompt (used for evaluation)}
]

\textcolor{qwenblue}{<|im\_start|>system}\\
{\color{textgray}\normalfont\itshape
You are an expert mathematical physicist solving a research-level problem. Produce a rigorous, self-contained solution using LaTeX notation. Address every numbered request, show the essential derivations, define conventions, and check signs, limits, or consistency where relevant. You have only the problem statement: do not assume or claim access to a reference answer. Do not use external tools, web sources, or unstated numerical experiments.
}
\\
\textcolor{qwenblue}{<|im\_end|>}

\vspace{1.5mm}

\textcolor{qwenblue}{<|im\_start|>user}\\
{\color{textgray}\normalfont
Solve the following problem completely. Preserve the problem's conventions.

\medskip

\texttt{<problem\_id>}\{problem\_id\}\texttt{</problem\_id>}

\medskip

\texttt{<problem\_latex>}

\{problem\}

\texttt{</problem\_latex>}
}
\\
\textcolor{qwenblue}{<|im\_end|>}

\vspace{1.5mm}

\textcolor{qwenblue}{<|im\_start|>assistant}\\
\textcolor{qwenblue}{<think>}\\
{\color{textgray}\normalfont\itshape
\{model-generated reasoning begins here; no reference completion is stored\}
}

\end{tcolorbox}

\clearpage
\section{Cross-Tokenizer Distillation Under a Unified View}
\label{sec:framework}

This section presents a unified view of cross-tokenizer on-policy distillation. We first define shared and non-shared token events and quantify direct vocabulary correspondence; we then characterize representative cross-tokenizer objectives by what they supervise and where supervision is applied; finally, we introduce Event-Set Completion Distillation (ESCD) and its training algorithm.

This section is organized as follows:

\begin{enumerate}
    \item \textbf{Basic definitions (\ref{sec:framework-definitions}):} we define byte realizations, shared and non-shared tokens, and the root and visited child states used throughout this section.

    \item \textbf{Direct vocabulary overlap (\ref{sec:framework-overlap}):} we quantify how much of the Teacher and Student vocabularies can be matched through direct one-to-one correspondence.

    \item \textbf{Existing methods under a unified view (\ref{sec:framework-existing-methods}):} we compare representative methods by asking what event they supervise and at which Student state that supervision is applied.

    \item \textbf{Event-Set Completion Distillation (\ref{sec:framework-escd}):} we define the residual completion set and the corresponding visited-child objective.
    
    \item \textbf{A worked example of cross-tokenizer supervision (\ref{sec:framework-worked-example}):} we compare SimCT, BPM, and ESCD using the same teacher and student probabilities, illustrating their target construction, loss calculations, and treatment of residual completion.
    
    \item \textbf{Training algorithm (\ref{sec:framework-training}):} we summarize how ESCD augments the standard on-policy pipeline and highlight the additional operations introduced by our method.
\end{enumerate}

\subsection{Basic Definitions}
\label{sec:framework-definitions}

\paragraph{On-policy predictions.}

Let \(T\) be a frozen Teacher and \(S_\theta\) a trainable Student with vocabularies \(V_T\) and \(V_S\). At a response prefix \(x\) visited by the Student rollout, they define next-token distributions

\begin{equation}
q_T(\cdot\mid x)\in\Delta(V_T),
\qquad
p_\theta(\cdot\mid x)\in\Delta(V_S).
\label{eq:framework-distributions}
\end{equation}

When the two models use the same tokenizer, these distributions can be compared directly at Student-visited states. Under tokenizer mismatch, however, token IDs are tokenizer-local coordinates and therefore do not define cross-model correspondence.

\paragraph{Shared and non-shared token events.}

Tokenizer-local IDs cannot be compared across models. We therefore describe each token in two different ways, each serving a distinct purpose.

First, let

\begin{equation}
b_T:V_T\rightarrow\mathcal B^*,
\qquad
b_S:V_S\rightarrow\mathcal B^*,
\label{eq:framework-byte-realization}
\end{equation}

where $b_T(v)$ and $b_S(u)$ are the exact byte strings produced by Teacher token $v$ and Student token $u$, respectively. These byte realizations are used throughout ESCD to define prefix relations, residual events, and completion sets.
Separately, we define tokenizer-independent matching keys

\begin{equation}
\kappa_T:V_T\rightarrow\mathcal K,
\qquad
\kappa_S:V_S\rightarrow\mathcal K.
\label{eq:framework-matching-key}
\end{equation}

The matching key is used only to identify tokens that can be transferred one-to-one across the two vocabularies. It removes a small number of explicitly specified tokenizer conventions, such as equivalent whitespace markers, and includes designated correspondences between semantically identical special tokens such as the primary EOS token. It is \emph{not} used to define ESCD byte events.

Using these keys, we construct a deterministic one-to-one shared-token mapping

\begin{equation}
\mathcal P_{\mathrm{sh}}
=
\operatorname{Select}_{1:1}
\left\{
(v,u)\in V_T\times V_S:
\kappa_T(v)=\kappa_S(u)
\right\}.
\label{eq:framework-shared-mapping}
\end{equation}

A Teacher token is \textbf{shared} if it appears in this mapping, and
\textbf{non-shared} otherwise:

\begin{equation}
V_{T,\mathrm{sh}}
=
\left\{
v\in V_T:
\exists u,\,(v,u)\in\mathcal P_{\mathrm{sh}}
\right\},
\qquad
V_{T,\mathrm{ns}}
=
V_T\setminus V_{T,\mathrm{sh}}.
\label{eq:framework-shared-sets}
\end{equation}

Shared tokens admit direct coordinate remapping through the selected pairs. Non-shared tokens are those outside this mapping; this designation does not necessarily imply the absence of an exact single-token byte counterpart. ESCD therefore uses a separate byte-based criterion to select child-event candidates, retaining content tokens whose complete byte strings have no single-token Student counterpart, as defined in Section~\ref{sec:completion_set}.

For example, Qwen3 and Qwen3.5 both contain the single-token string \texttt{['apple']}, although their local token IDs differ. These tokens form a shared pair. In contrast, Qwen3 token 382 realizes \texttt{['.\textbackslash n\textbackslash n']}, whereas Qwen3.5 realizes the same bytes with the two-token sequence \texttt{[13, 271]}. This example illustrates how the same bytes can be realized through different token boundaries.

\paragraph{Root and visited child states.}

Let \(x\) denote the full response prefix generated so far by the Student rollout, i.e., the context on which the next-token prediction is conditioned. We call \(x\) the \emph{root state}, as it serves as the common starting point for the continuation paths considered below. At this state, consider a non-shared Teacher token \(v\) whose byte realization is

\begin{equation}
y=b_T(v).
\label{eq:framework-teacher-event}
\end{equation}

The Student then samples its next token $a\sim p_\theta(\cdot\mid x)$. If the bytes produced by this actually sampled token form a strict prefix of the Teacher event,

\begin{equation}
b_S(a)\prec y,
\label{eq:framework-prefix-condition}
\end{equation}

then the rollout naturally reaches the next state after appending $a$ to $x$. We denote this \emph{visited child state} by $(x,a)$. The part of the Teacher event that remains unfinished at this state is

\begin{equation}
r=y[|b_S(a)|:].
\label{eq:framework-residual}
\end{equation}

For example, suppose the Teacher assigns probability to a non-shared event whose byte surface is \texttt{Bread}. At the current Student-generated prefix $x$, the Student actually emits the token \texttt{Br}. Then

\begin{equation}
\underbrace{x}_{\text{root state}}
\;\xrightarrow{\;a=\texttt{Br}\;}
\underbrace{(x,\texttt{Br})}_{\text{visited child state}},
\qquad
\underbrace{y=\texttt{Bread}}_{\text{Teacher event}}
=
\underbrace{\texttt{Br}}_{b_S(a)}
\Vert
\underbrace{\texttt{ead}}_{r}.
\label{eq:framework-bread-example}
\end{equation}

Here, $x$ is the entire Student-generated response prefix before the current action, rather than a single token. The variable $a$ is the Student token actually selected at that state, and $r$ is the remaining part of the Teacher byte event after removing the bytes already realized by $a$.

This distinction is central to ESCD. The Teacher event is initially defined at the root state $x$. Once the Student naturally takes a compatible action $a$, ESCD can continue supervising the remaining event $r$ at the visited child state $(x,a)$, without constructing a counterfactual Student branch.

For the remainder of this section, we compare methods along two questions: \emph{how is a non-shared Teacher event represented for the Student, and at which Student state is it supervised?}

\subsection{How Much of the Vocabulary Is Directly Shared?}
\label{sec:framework-overlap}

Before considering non-shared events, we quantify how much direct one-to-one correspondence already exists between the two vocabularies. Let

\begin{equation}
M
=
|\mathcal P_{\mathrm{sh}}|
\label{eq:framework-shared-count}
\end{equation}

denote the number of selected shared pairs. We report Teacher-side coverage, Student-side coverage, and Jaccard overlap:

\begin{figure*}[t]
    \centering
    \includegraphics[width=\textwidth]
    {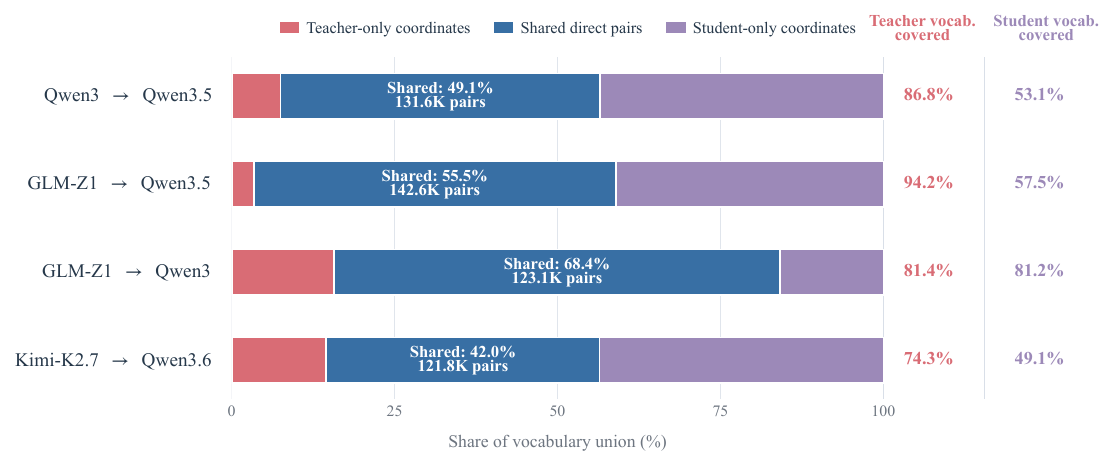}
    \vspace{-6mm}
\caption{Static Teacher--Student vocabulary overlap. Bars show Teacher-only, shared, and Student-only shares of the vocabulary union; right-hand columns report directional coverage.}
    \label{fig:teacher-student-vocabulary-overlap}
\end{figure*}

\begin{equation}
C_T
=
\frac{M}{|V_T|},
\qquad
C_S
=
\frac{M}{|V_S|},
\qquad
J
=
\frac{M}{|V_T|+|V_S|-M}.
\label{eq:framework-overlap-metrics}
\end{equation}

Here $C_T$ measures the fraction of Teacher coordinates that can be transferred directly, $C_S$ gives the corresponding Student-side coverage, and $J$ measures overlap relative to the vocabulary union.

\begin{table}[t]
\centering
\caption{Static shared-vocabulary overlap under the deterministic one-to-one canonical-surface mapping. Vocabulary sizes exclude checkpoint padding.}
\label{tab:framework-static-overlap}
\begin{tabular}{lcccccc}
\toprule
\textbf{Tokenizer pair}
& $\mathbf{|V_T|}$
& $\mathbf{|V_S|}$
& $\mathbf{M}$
& $\mathbf{C_T}$
& $\mathbf{C_S}$
& $\mathbf{J}$ \\
\midrule
Qwen3 $\rightarrow$ Qwen3.5
& 151,669 & 248,077 & 131,612
& 86.78\% & 53.05\% & 49.08\% \\

Qwen3.5 $\rightarrow$ Qwen3
& 248,077 & 151,669 & 131,612
& 53.05\% & 86.78\% & 49.08\% \\

GLM-Z1 $\rightarrow$ Qwen3.5
& 151,343 & 248,077 & 142,628
& 94.24\% & 57.49\% & 55.54\% \\

GLM-Z1 $\rightarrow$ Qwen3
& 151,343 & 151,669 & 123,118
& 81.35\% & 81.18\% & 68.44\% \\

Kimi-K2.7 $\rightarrow$ Qwen3.6
& 163,840 & 248,077 & 121,757
& 74.31\% & 49.08\% & 41.96\% \\

\bottomrule
\end{tabular}
\end{table}

Table~\ref{tab:framework-static-overlap} shows that shared coordinates cover 53.05--94.24\% of the Teacher vocabulary across evaluated directions. Direct remapping handles these coordinates, while our method targets the substantial non-shared subset requiring explicit cross-tokenizer treatment.

\subsection{Existing Methods Under the Unified View}
\label{sec:framework-existing-methods}

We compare representative methods by their cross-tokenizer alignment mechanisms and supervision objects, including ranked probability profiles, mapped vocabulary coordinates, and aligned continuation units. This distinguishes how methods construct comparable predictions from how they transfer supervision. Because position- and span-alignment conventions differ, prediction vectors need not share an identical conditioning prefix. The formulations below summarize the relevant distillation components and their relation to ESCD's completion-set supervision.

\paragraph{ULD: rank-based distribution matching.}

ULD compares probability profiles without constructing token-identity correspondences~\citep{uld}. Let $q_i$ and $p_i$ denote the Teacher and Student probability vectors at the positions being compared. After padding the smaller vocabulary distribution with zeros to length $K=\max(|V_T|,|V_S|)$, define

\begin{equation}
q_i^{\downarrow}(k)
=
\text{$k$-th largest entry of the padded }q_i,
\qquad
p_i^{\downarrow}(k)
=
\text{$k$-th largest entry of the padded }p_i.
\label{eq:framework-uld-sort}
\end{equation}

The rank-matching term is

\begin{equation}
\ell_{\mathrm{ULD}}(i)
=
\sum_{k=1}^{K}
\left|
q_i^{\downarrow}(k)-p_i^{\downarrow}(k)
\right|.
\label{eq:framework-uld}
\end{equation}

Rank $k$ identifies a probability magnitude rather than a shared token, so this term transfers distributional shape without preserving textual correspondence between coordinates. The original training objective also includes cross-entropy supervision and compares sequence positions up to the shorter tokenized length. Rank matching itself does not resolve mismatched text boundaries; applying it after an additional alignment procedure is a separate implementation choice.

\paragraph{GOLD: span alignment and hybrid matching.}

GOLD combines cross-tokenizer sequence alignment and probability merging with configurable distribution-matching losses. In its hybrid formulation, directly matched coordinates retain token identity, while unmatched coordinates are compared through sorted probability profiles. Let $\widehat q_T^{(k)}$ and $\widehat p_\theta^{(k)}$ denote the Teacher and Student probability representations associated with aligned unit $k$ after merging. A schematic hybrid loss is

\begin{equation}
\ell_{\mathrm{GOLD}}(k)
=
\lambda_{\mathrm{sh}}
D_{\mathrm{sh}}
\left(
\widehat q_{T,\mathrm{sh}}^{(k)},
\widehat p_{\theta,\mathrm{sh}}^{(k)}
\right)
+
\lambda_{\mathrm{ns}}
\left\|
\operatorname{sort}_{\downarrow}
\widehat q_{T,\mathrm{ns}}^{(k)}
-
\operatorname{sort}_{\downarrow}
\widehat p_{\theta,\mathrm{ns}}^{(k)}
\right\|_1,
\label{eq:framework-gold}
\end{equation}
where $\operatorname{sort}_{\downarrow}$ sorts entries in descending order, $\|\mathbf z\|_1=\sum_j |z_j|$ denotes the $L_1$ norm, and unmatched vectors are zero-padded to equal dimension before comparison. Here, $D_{\mathrm{sh}}$ denotes the configured shared-coordinate discrepancy, and $\lambda_{\mathrm{sh}}$ and $\lambda_{\mathrm{ns}}$ weight the two terms. The shared term preserves explicit coordinate correspondences, whereas the unmatched term compares probabilities by rank rather than token identity.
Sequence merging can incorporate conditional probabilities from multiple token positions, so GOLD operates on aligned, merged representations rather than necessarily comparing raw next-token distributions at a common boundary. Its hybrid matching objective is distinct from an auxiliary loss on the valid completions of a specified residual byte event.

\paragraph{X-Token: aligned-chunk vocabulary projection.}

X-Token combines span alignment, chain-rule probability merging, and vocabulary projection~\citep{sreenivas2026xtokenprojectionguidedcrosstokenizerknowledge}. Aligned spans provide text-consistent comparison units, while a sparse matrix $W\in\mathbb R^{|V_S|\times|V_T|}$ maps Student coordinates into Teacher vocabulary space. The mapping is initialized from canonicalized token matches and multi-token decompositions, with row normalization; it can optionally be refined during training.

For the P-KL formulation, let $\widehat p_\theta^{(k)}$ and $\widehat q_T^{(k)}$ denote the aligned chunk distributions. The projected Student distribution is

\begin{equation}
\widetilde p_{\theta,W}^{(k)}(v)
=
\sum_{u\in V_S}
W_{u,v}\widehat p_\theta^{(k)}(u),
\qquad v\in V_T.
\label{eq:framework-xtoken-projection}
\end{equation}

The corresponding distillation term is

\begin{equation}
\ell_{\mathrm{XToken\text{-}P}}(k)
=
\operatorname{KL}
\left(
\widehat q_T^{(k)}
\,\middle\|\,
\widetilde p_{\theta,W}^{(k)}
\right).
\label{eq:framework-xtoken}
\end{equation}

X-Token also introduces H-KL, which uses high-confidence mappings to expand the matched set within a hybrid objective. Thus, the projection above describes P-KL rather than every X-Token variant. Although the initialization of $W$ depends on tokenizer structure, the complete method also uses sequence-level alignment and merging. Its central operation is distribution matching over aligned chunks, rather than constructing a completion set for an unfinished Teacher event.

\paragraph{SimCT: scoring minimal aligned continuation units.}

SimCT constructs a common supervision space of shared tokens and minimal aligned units realizable by both tokenizers~\citep{simct}. These units may have different token counts in each model. For $M\in\{T,S\}$, let $\tau_M(c)=(v_1,\ldots,v_{L_M(c)})$ denote the tokenization of candidate unit $c$. SimCT computes length-normalized continuation scores and normalizes them over the candidate space $\mathcal U(x)$:

\begin{equation}
\begin{aligned}
s_M(c\mid x)
&=
\frac{1}{L_M(c)}
\sum_{j=1}^{L_M(c)}
\log p_M(v_j\mid x,v_{<j}),
\\
\pi_M(c\mid x)
&=
\frac{\exp s_M(c\mid x)}
{\sum_{c'\in\mathcal U(x)}\exp s_M(c'\mid x)}.
\end{aligned}
\label{eq:framework-simct-scores}
\end{equation}

Here $p_M$ denotes the native autoregressive probabilities of either model. An OPD divergence is then applied to the induced distributions:

\begin{equation}
\mathcal L_{\mathrm{SimCT}}(x)
=
D_{\mathrm{OPD}}
\left(
\pi_S(\cdot\mid x),
\pi_T(\cdot\mid x)
\right).
\label{eq:framework-simct}
\end{equation}

These are normalized continuation-score distributions, not mass-preserving marginals of the original next-token distributions. Candidate continuations are scored through autoregressive factors from the conditioning prefix; their intermediate states need not all be visited by the actual rollout. 

\paragraph{BPM: Teacher-induced byte-prefix targets.}

BPM maps Teacher probability mass into Student-token targets through byte-prefix relations~\citep{bpm}. In the basic refinement case, define the longest compatible Student-token prefix of Teacher content token $v$:

\begin{equation}
\phi(v)
=
\operatorname*{arg\,max}_{u\in V_S:\,b_S(u)\preceq b_T(v)}
|b_S(u)|,
\label{eq:framework-bpm-map}
\end{equation}

with $\phi(v)=\bot$ when no eligible prefix exists. At aligned position $i$, the target aggregates Teacher probabilities:

\begin{equation}
t_i(u)
=
\sum_{v:\phi(v)=u}q_i(v),
\qquad
t_i(\varnothing)
=
\sum_{v:\phi(v)=\bot}q_i(v).
\label{eq:framework-bpm}
\end{equation}

BPM also supervises positions inside a Teacher token. If $c$ denotes the bytes already emitted since the Teacher boundary, let $\phi_c(v)$ map the remaining bytes to their longest Student-token prefix. For $\pi_i(c)=\sum_{v:\,c\preceq b_T(v)}q_i(v)>0$, the conditional target is

\begin{equation}
t_i(u\mid c)
=
\frac{
\sum_{v:\,c\preceq b_T(v),\,\phi_c(v)=u}q_i(v)
}{
\pi_i(c)
},
\qquad
t_i(\varnothing\mid c)
=
1-\sum_{u\in V_S}t_i(u\mid c).
\label{eq:framework-bpm-conditional}
\end{equation}

The original method additionally handles Student tokens spanning Teacher boundaries and model-specific stopping events. Its targets are matched to Student probabilities through a distributional distillation loss. BPM therefore cannot be characterized as root-only supervision or as summing probabilities over all Student tokenization paths. The relevant distinction is that ESCD adds a Teacher-mass-weighted loss on the aggregate probability of valid residual completions, without assigning an individual target probability to each completion.

\paragraph{Relation to ESCD.}

These methods establish supervision through ranked coordinates, matched tokens, aligned spans, or Teacher-induced token targets. ESCD introduces a complementary supervision unit: the set of native Student actions that complete a residual byte constraint at a naturally visited child state. Existing distribution-matching objectives can also constrain aggregate probabilities indirectly; ESCD makes this particular completion event explicit through an auxiliary set-probability loss. Its distinction therefore concerns the construction and granularity of the completion target, rather than the presence of supervision at later Student positions alone.

\begin{table*}[hbt]
\centering
\caption{Cross-tokenizer supervision mechanisms. Entries summarize distillation components, not full training recipes. ESCD explicitly supervises residual completion sets.}
\label{tab:framework-method-comparison}
\small
\setlength{\tabcolsep}{5pt}
\renewcommand{\arraystretch}{1.15}
\begin{adjustbox}{max width=\textwidth}
\begin{tabular}{@{}lll@{}}
\toprule
\textbf{Method}
& \textbf{Alignment or mapping mechanism}
& \textbf{Supervision form} \\
\midrule
ULD
& Independent probability sorting
& Rank-matched probability profiles \\
GOLD
& Span alignment and probability merging
& Shared-coordinate and unmatched-profile matching \\
X-Token
& Span alignment and vocabulary projection
& Projected or hybrid chunk-level distribution matching \\
SimCT
& Minimal aligned continuation units
& Normalized continuation-score distribution matching \\
BPM
& Byte-prefix mapping and conditional targets
& Teacher-induced Student-token distribution matching \\
\rowcolor{OPDBlue!7}
\textbf{ESCD}
& \textbf{Prefix aggregation and visited residual construction}
& \textbf{Auxiliary loss on completion-set probability} \\
\bottomrule
\end{tabular}
\end{adjustbox}
\end{table*}

\subsection{ESCD: Event-Set Completion Distillation}
\label{sec:framework-escd}

\paragraph{From event entry to event completion.}

A Student action may realize only a prefix of a Teacher byte event, leaving a residual constraint at the next visited state. Existing cross-tokenizer methods can supervise intermediate Student positions; ESCD instead makes the residual completion set an explicit supervision object. It combines BPM-style root projection with a loss on the aggregate probability of valid completions at visited child states.

Consider an illustrative Teacher token with byte realization \texttt{Bread}, which the Student can realize through \texttt{Br} followed by \texttt{ead}:

\begin{equation}
\underbrace{\texttt{Bread}}_{\text{Teacher event}}
\qquad\Longleftrightarrow\qquad
\underbrace{\texttt{Br}}_{\text{Student action }a}
\mathbin{\Vert}
\underbrace{\texttt{ead}}_{\text{residual bytes}},
\label{eq:framework-escd-entry-example}
\end{equation}

where $\mathbin{\Vert}$ denotes byte-string concatenation. After sampling $a=\texttt{Br}$ at state $x$, the Student reaches

\begin{equation}
x\xrightarrow{\;\texttt{Br}\;}x'=(x,\texttt{Br}),
\qquad
\texttt{Bread}=\texttt{Br}\mathbin{\Vert}\texttt{ead}.
\label{eq:framework-escd-residual-example}
\end{equation}

At $x'$, ESCD supervises the total probability of next-token actions that complete the remaining bytes, potentially extending beyond the event boundary. It neither selects a canonical completion nor assigns individual target probabilities to valid actions.

\paragraph{Root projection loss.}

ESCD reuses BPM's byte alignment and root projection, but not its full interior- and spanning-position supervision. At an eligible aligned root $x$, let $\mathcal K_T(x)$ denote the selected Teacher candidates, $U(x)\subseteq V_S$ the explicit Student-token coordinates, and $\phi_x$ the root byte mapping, with $\phi_x(v)=\bot$ for candidates not assigned to an explicit coordinate. The mapped Teacher mass is

\begin{equation}
q_{\mathrm{map}}(u\mid x)
=
\sum_{\substack{v\in\mathcal K_T(x)\\ \phi_x(v)=u}}
q_T(v\mid x),
\qquad u\in U(x).
\label{eq:framework-escd-root-mass}
\end{equation}

We construct Teacher and Student distributions over the same coordinates $U(x)\cup\{\bot\}$:

\begin{equation}
\begin{aligned}
\bar q_x(u)
&=q_{\mathrm{map}}(u\mid x),
&
\bar p_{\theta,x}(u)
&=p_\theta(u\mid x),
\quad u\in U(x),
\\
\bar q_x(\bot)
&=1-\sum_{u\in U(x)}q_{\mathrm{map}}(u\mid x),
&
\bar p_{\theta,x}(\bot)
&=1-\sum_{u\in U(x)}p_\theta(u\mid x).
\end{aligned}
\label{eq:framework-escd-root-projection}
\end{equation}

The complement $\bot$ collects mass outside the explicit coordinates and is not a generatable token. Teacher probabilities retain their original values after candidate selection; omitted and unmapped mass remains in the complement. We apply forward KL:

\begin{equation}
\ell_{\mathrm{root}}(x)
=
\operatorname{KL}\!\left(
\bar q_x \,\middle\|\, \bar p_{\theta,x}
\right).
\label{eq:framework-escd-root-loss}
\end{equation}

\paragraph{Teacher-side prefix aggregation.}

Child candidates are selected by exact byte realizability. Let $V_T^{\mathrm{cont}}$ and $V_S^{\mathrm{cont}}$ denote the Teacher and Student content-token vocabularies. We retain Teacher candidates whose complete byte strings have no single-token Student counterpart:

\begin{equation}
\mathcal V(x)
=
\left\{
v\in\mathcal K_T(x)\cap V_T^{\mathrm{cont}}
:
\nexists u\in V_S^{\mathrm{cont}},
\ b_T(v)=b_S(u)
\right\}.
\label{eq:framework-escd-child-candidates}
\end{equation}

This eligibility rule excludes special tokens and exact single-token byte matches. It is distinct from merely excluding tokens selected by the one-to-one shared-coordinate mapping.

Distinct Teacher tokens are mutually exclusive outcomes, but their byte-prefix constraints may be nested. For example, satisfying the prefix \texttt{researches} also satisfies \texttt{resear}. ESCD aggregates such constraints into a representative prefix. We connect $v,w\in\mathcal V(x)$ whenever $b_T(v)\preceq b_T(w)$ or $b_T(w)\preceq b_T(v)$ and use the connected components as aggregation groups. For group $g$ with members $\mathcal V_g$, define

\begin{equation}
M_g
=
\sum_{v\in\mathcal V_g}q_T(v\mid x).
\label{eq:framework-escd-component-mass}
\end{equation}

The representative is a shortest byte realization in the component:

\begin{equation}
v_g^\star
\in
\operatorname*{arg\,min}_{v\in\mathcal V_g}|b_T(v)|,
\qquad
y_g=b_T(v_g^\star),
\label{eq:framework-escd-component-event}
\end{equation}

with deterministic tie breaking. The representative $y_g$ prefixes every member, and representatives of distinct groups are prefix-incomparable. Aggregation preserves the included Teacher mass while retaining only the shared representative constraint, not each longer member's full byte requirement.

\paragraph{Completion sets at visited Student states.}

Suppose the Student samples a content token $a$ whose bytes form a nonempty strict prefix of $y_g$:

\begin{equation}
0<|b_S(a)|<|y_g|,
\qquad
b_S(a)\prec y_g.
\label{eq:framework-escd-prefix}
\end{equation}

At the visited child state $x'=(x,a)$, the residual is

\begin{equation}
r_g=y_g[|b_S(a)|:].
\label{eq:framework-escd-residual}
\end{equation}

The valid one-step completion set is

\begin{equation}
\mathcal C(r_g)
=
\left\{
u\in V_S^{\mathrm{cont}}
:
r_g\preceq b_S(u)
\right\}.
\label{eq:framework-completion-set}
\end{equation}

A valid token begins with the entire residual and may extend beyond its endpoint. For $r_g=\texttt{Memory}$, both \texttt{Memory} and \texttt{MemoryWarning} qualify if present in the Student vocabulary. The latter satisfies the residual constraint without implying Teacher supervision on the additional suffix \texttt{Warning}.
The completion probability sums native full-vocabulary probabilities:

\begin{equation}
P_\theta(\mathcal C(r_g)\mid x,a)
=
\sum_{u\in\mathcal C(r_g)}
p_\theta(u\mid x,a).
\label{eq:framework-escd-completion-mass}
\end{equation}

All probabilities are evaluated at the same visited child state, without renormalization over content tokens or the completion set. No alternative continuations are sampled, and the actual next token need not belong to $\mathcal C(r_g)$ for the loss to apply.

\paragraph{Local child multiplicity.}

For a Teacher group $g$, suppose the sampled Student action $a$ satisfies $b_S(a)\prec y_g$, leaving part of the representative event unfinished. To distinguish Teacher-side aggregation from Student-side completion, define
\begin{equation}
m_g=|V_g|,
\qquad
n_g(a)=|\mathcal C(r_g)|,
\qquad
r_g=y_g[|b_S(a)|:].
\label{eq:framework-escd-multiplicity}
\end{equation}
Here, $m_g$ counts Teacher candidates in the group, while $n_g(a)$ counts Student tokens that complete the residual in one additional step. The latter depends on the sampled action and may be zero; only groups with $n_g(a)>0$ contribute to the child loss. These quantities describe local supervision structure, not token counts in an aligned span. In particular, $n_g(a)$ differs from the root-compatible token count $b_g$ used in Table~\ref{tab:geometry-full-trajectory}.

The following examples illustrate different combinations of Teacher group size and Student completion count, assuming eligible Teacher candidates and exactly the listed completion sets.
\begin{itemize}

\item \textbf{Single-member group, single completion ($m_g=1$, $n_g(a)=1$).}
For $y_g=\texttt{Bread}$ and $a=\texttt{Br}$, the residual is \texttt{ead}. If $V_g$ contains one Teacher token and $\mathcal C(r_g)=\{\texttt{ead}\}$, the child loss reduces to $-M_g\log p_\theta(\texttt{ead}\mid x,a)$.

\item \textbf{Single-member group, multiple completions ($m_g=1$, $n_g(a)>1$).}
For $y_g=\texttt{(Memory}$ and $a=\texttt{(}$, the residual is \texttt{Memory}. Suppose $V_g$ contains one Teacher token and $\mathcal C(r_g)=\{\texttt{Memory},\texttt{MemoryWarning}\}$. The child loss is
\begin{equation}
-M_g\log\left[
p_\theta(\texttt{Memory}\mid x,a)
+
p_\theta(\texttt{MemoryWarning}\mid x,a)
\right].
\label{eq:framework-escd-one-to-many-example}
\end{equation}

\item \textbf{Multi-member group, single completion ($m_g>1$, $n_g(a)=1$).}
Suppose a group consists of \texttt{[space]resear} and \texttt{[space]researches}, where \texttt{[space]} denotes a literal leading space. Its mass is
\begin{equation}
M_g
=
q_T(\texttt{[space]resear}\mid x)
+
q_T(\texttt{[space]researches}\mid x).
\label{eq:framework-escd-many-to-one-example}
\end{equation}
After the Student emits the space, the representative residual is \texttt{resear}. If $\mathcal C(r_g)=\{\texttt{research}\}$, then $m_g=2$ and $n_g(a)=1$. This completion satisfies the representative constraint but leaves the longer event \texttt{researches} unfinished, illustrating the coarsening introduced by aggregation.

\item \textbf{Multi-member group, multiple completions ($m_g>1$, $n_g(a)>1$).}
Suppose a group consists of \texttt{VEH} and \texttt{VEHICLE}, with representative \texttt{VEH}. After $a=\texttt{V}$, the residual is \texttt{EH}. If $\mathcal C(r_g)=\{\texttt{EH},\texttt{EHICLE}\}$, then $m_g=2$ and $n_g(a)=2$: Teacher mass is aggregated across group members, and Student probability is summed across valid completions.

\end{itemize}

\paragraph{Child objective.}

Let $\mathcal G(x,a)$ contain groups for which the sampled action $a$ partially realizes the representative event, as specified in Equation~\ref{eq:framework-escd-prefix}, and the residual admits at least one valid one-step completion ($n_g(a)>0$). At the visited child state, the loss is
\begin{equation}
\ell_{\mathrm{child}}(x,a)
=
-\sum_{g\in\mathcal G(x,a)}
M_g
\log P_\theta(\mathcal C(r_g)\mid x,a).
\label{eq:framework-escd-child}
\end{equation}

The weights $M_g$ retain their root Teacher mass without renormalization over eligible groups. They weight residual constraints rather than define a Teacher posterior conditioned on $a$. No additional Teacher query is required. The loss is zero when $\mathcal G(x,a)$ is empty.

\paragraph{Combined objective.}

Root terms are assigned to their aligned prediction positions, and child terms to the immediately following visited prediction positions. Their per-position losses are added before applying the training mask and reduction. Writing $\mathcal L_{\mathrm{root}}$ and $\mathcal L_{\mathrm{child}}$ for the two contributions under this common reduction, the full objective is

\begin{equation}
\mathcal L_{\mathrm{ESCD}}
=
\mathcal L_{\mathrm{root}}
+
\mathcal L_{\mathrm{child}}.
\label{eq:framework-escd-full}
\end{equation}

Both terms use the same reduction denominator, with zero contributions at ineligible positions; they are not separately averaged over their eligible positions. ESCD thus combines root projection and child completion rather than adding the child term to the complete BPM objective.

\paragraph{Why supervise a completion set?}

Selecting one valid completion introduces a preference within the set. For any $u^\star\in\mathcal C(r_g)$, the single-completion loss decomposes as

\begin{equation}
\begin{aligned}
-M_g\log p_\theta(u^\star\mid x,a)
={}&
-M_g\log P_\theta(\mathcal C(r_g)\mid x,a)
\\
&-
M_g\log
\frac{
p_\theta(u^\star\mid x,a)
}{
P_\theta(\mathcal C(r_g)\mid x,a)
}.
\end{aligned}
\label{eq:framework-escd-single-decomposition}
\end{equation}

The first term encourages residual completion, while the second favors the selected token within the completion set. ESCD uses only the first term, without specifying a within-set target distribution. The objectives coincide when $n_g(a)=1$; when $n_g(a)>1$, the single-completion loss introduces an additional preference not specified by the residual constraint.

\paragraph{Scope.}

ESCD supervises a representative residual byte constraint, not the full byte requirement of every original Teacher token. Its current child objective covers completion by one additional Student token; residuals without a valid one-step completion contribute no child loss. Completion probabilities reuse Student logits at visited states, without canonical continuations or counterfactual rollouts.
Teacher candidates may be selected from the full vocabulary or a sparse subset. With sparse selection, groups, representatives, and weights are constructed from the retained candidates using their original probabilities. Candidate restriction can therefore change the supervision itself, not merely its computational cost.

\subsection{A Worked Example of Cross-Tokenizer Supervision}
\label{sec:framework-worked-example}

The preceding sections characterize the supervision mechanisms of SimCT, BPM, and ESCD. To make their differences concrete, we work through a numerical example in which teacher tokens are realized by multiple student tokens. Using the same hypothetical predictions, we trace how each method constructs its target and computes its loss, clarifying how subsequent student actions enter the objective and whether individual completions are distinguished. All probabilities are illustrative, logarithms are natural, and losses are reported before training-mask reduction.

A natural question is: if \texttt{Bread} receives the highest teacher probability, why do BPM and ESCD still retain other teacher candidates such as \texttt{Breads} and \texttt{Break}? Does including lower-probability candidates introduce undesirable supervision? The answer is no, because these candidates are alternative next-token predictions from the teacher distribution, rather than correctness labels. Their probabilities represent the teacher's preference over possible next-token events under the same context. After the student samples \texttt{Br}, BPM and ESCD preserve this teacher information at different granularities. BPM maintains token-level distinctions and assigns separate targets to the residual tokens \texttt{ead}, \texttt{eads}, and \texttt{eak}. ESCD instead aggregates teacher mass over residual byte constraints: \texttt{Bread} and \texttt{Breads} are grouped because they share the same representative prefix constraint, while \texttt{Break} remains a separate event with its own teacher mass. Therefore, ESCD does not treat \texttt{Breads} as equivalent to \texttt{Bread}; it only removes the need to allocate probability among tokenizer-dependent realizations that satisfy the same residual constraint.

Table~\ref{tab:framework-worked-example} gives the teacher distribution at root context $x$. For simplicity, the four candidates carry all teacher probability mass. Assume the student has no single-token representation of \texttt{Bread}, \texttt{Breads}, or \texttt{Break}, and \texttt{Br} is their longest student-token prefix. The stated student tokenizations are assumed throughout; \texttt{Cat} is shared directly.

\begin{table}[hbt]
\centering
\small
\setlength{\tabcolsep}{7pt}
\caption{Illustrative teacher predictions and student realizations. Residuals are shown for individual teacher tokens after \texttt{Br}, before ESCD aggregation; \texttt{Cat} is incompatible with this action.}
\label{tab:framework-worked-example}
\begin{tabular}{@{}lclc@{}}
\toprule
Teacher token
& $q_T(v\mid x)$
& Student tokenization
& Residual after \texttt{Br} \\
\midrule
\texttt{Bread}  & $0.4$ & \texttt{Br}, \texttt{ead}  & \texttt{ead} \\
\texttt{Breads} & $0.2$ & \texttt{Br}, \texttt{eads} & \texttt{eads} \\
\texttt{Break}  & $0.3$ & \texttt{Br}, \texttt{eak}  & \texttt{eak} \\
\texttt{Cat}    & $0.1$ & \texttt{Cat}               & --- \\
\bottomrule
\end{tabular}
\end{table}

At the root, let
\begin{equation}
p_\theta(\texttt{Br}\mid x)=0.5,
\qquad
p_\theta(\texttt{Cat}\mid x)=0.2,
\end{equation}
with probability $0.3$ assigned to all remaining tokens. Suppose the student actually samples $a=\texttt{Br}$ and reaches $x'=(x,a)$. Its next-token distribution at this visited child is
\begin{equation}
\begin{aligned}
p_\theta(\texttt{ead}\mid x') &= 0.2, &
p_\theta(\texttt{eads}\mid x') &= 0.3,\\
p_\theta(\texttt{eak}\mid x') &= 0.1, &
\sum_{u\notin\{\texttt{ead},\texttt{eads},\texttt{eak}\}}
p_\theta(u\mid x') &= 0.4.
\end{aligned}
\label{eq:worked-example-student}
\end{equation}
These are native full-vocabulary probabilities, without renormalization over the displayed tokens.

\textbf{SimCT} compares distributions over aligned continuation units. Here, the candidate space is
\begin{equation}
\mathcal U
=
\{\texttt{Bread},\texttt{Breads},\texttt{Break},\texttt{Cat}\}.
\end{equation}
Each teacher realization contains one token, whereas the first three student realizations contain two. SimCT averages the log-probabilities along each realization. For \texttt{Bread},
\begin{equation}
\begin{aligned}
s_T(\texttt{Bread}\mid x)
&=\log0.4,\\
s_S(\texttt{Bread}\mid x)
&=\frac12\left[\log0.5+\log0.2\right].
\end{aligned}
\end{equation}
The path probability is $0.5\times0.2=0.1$, but the exponentiated, length-normalized score is $\sqrt{0.1}$.

In the listed order, the exponentiated scores are
\begin{equation}
\begin{aligned}
w_T&=(0.4,0.2,0.3,0.1),\\
w_S&=(\sqrt{0.1},\sqrt{0.15},\sqrt{0.05},0.2).
\end{aligned}
\end{equation}
The scores sum to $1$ for the teacher and approximately $1.1271$ for the student. Normalizing each vector over $\mathcal U$ yields
\begin{equation}
\begin{aligned}
\pi_T&=(0.4,0.2,0.3,0.1),\\
\pi_S&\approx(0.2806,0.3436,0.1984,0.1774).
\end{aligned}
\end{equation}
The reverse-KL formulation then gives
\begin{equation}
\mathcal L_{\mathrm{SimCT}}
=
\operatorname{KL}(\pi_S\|\pi_T)
\approx0.1062.
\end{equation}

SimCT supervises subsequent predictions through complete-unit scores, keeping \texttt{Bread} and \texttt{Breads} as separate coordinates. However, candidate normalization can obscure changes in native completion probability. To see this, keep the teacher distribution and $\mathcal U$ fixed and halve the displayed student probabilities at both $x$ and $x'$, reallocating the removed mass to other tokens. The root probabilities of \texttt{Br} and \texttt{Cat} become $(0.25,0.1)$, and the child probabilities of \texttt{ead}, \texttt{eads}, and \texttt{eak} become $(0.1,0.15,0.05)$. Every exponentiated unit score is then halved, leaving the normalized distribution and local SimCT loss unchanged. Yet the probability of completing \texttt{ead} after \texttt{Br} falls from $0.5$ to $0.25$. Thus, this objective on the fixed candidate space does not uniquely determine completion probability at $x'$.

\textbf{BPM} constructs teacher-induced student-token distributions. We use its basic refinement targets with forward KL; spanning-token corrections and stopping cases are outside this example. At the root, the longest-prefix map sends \texttt{Bread}, \texttt{Breads}, and \texttt{Break} to \texttt{Br}, giving
\begin{equation}
t_x(\texttt{Br})=0.4+0.2+0.3=0.9,
\qquad
t_x(\texttt{Cat})=0.1.
\end{equation}

Over coordinates $(\texttt{Br},\texttt{Cat},\bot)$, where $\bot$ collects the remaining mass, the projected distributions and root loss are
\begin{equation}
t_x=(0.9,0.1,0),
\qquad
\bar p_{\theta,x}=(0.5,0.2,0.3).
\end{equation}
\begin{equation}
\ell_{\mathrm{BPM,root}}
=\operatorname{KL}(t_x\|\bar p_{\theta,x})
=0.9\log\frac{0.9}{0.5}+0.1\log\frac{0.1}{0.2}
\approx0.4597.
\label{eq:worked-example-bpm-root}
\end{equation}
The complement is an aggregate coordinate, not a generatable token; neither distribution is renormalized over only \texttt{Br} and \texttt{Cat}.

After \texttt{Br}, the compatible teacher mass is $0.4+0.2+0.3=0.9$. BPM conditions these candidates on the emitted prefix and maps each residual to its longest student-token prefix:
\begin{equation}
\begin{aligned}
\texttt{Bread}&\longmapsto\texttt{ead},
&t_{x'}(\texttt{ead})&=0.4/0.9=4/9,\\
\texttt{Breads}&\longmapsto\texttt{eads},
&t_{x'}(\texttt{eads})&=0.2/0.9=2/9,\\
\texttt{Break}&\longmapsto\texttt{eak},
&t_{x'}(\texttt{eak})&=0.3/0.9=3/9.
\end{aligned}
\end{equation}
Here $t_{x'}$ denotes the conditional target derived from the root teacher prediction, rather than a new teacher query at $x'$. The incompatible \texttt{Cat} candidate is excluded. The denominator $0.9$ is teacher mass, not the student's action probability $0.5$.

In the coordinate order $(\texttt{ead},\texttt{eads},\texttt{eak},\bot)$, the distributions are $t_{x'}=(4/9,2/9,3/9,0)$ and $\bar p_{\theta,x'}=(0.2,0.3,0.1,0.4)$. Hence
\begin{equation}
\ell_{\mathrm{BPM,interior}}
=\frac49\log\frac{4/9}{0.2}
+\frac29\log\frac{2/9}{0.3}
+\frac39\log\frac{3/9}{0.1}
\approx0.6895.
\label{eq:worked-example-bpm-interior}
\end{equation}
The root and interior losses apply at distinct prediction positions and sum to approximately $1.1492$ before the common response-token reduction. They are not averaged separately by position type.

For the representative residual \texttt{ead} used by ESCD below, both \texttt{ead} and \texttt{eads} are valid one-step completions. They remain distinct full continuations, however, and BPM preserves the teacher's $2{:}1$ preference. Keeping the root predictions fixed, changing their child probabilities from $(0.2,0.3)$ to $(0.1,0.5)$, with \texttt{eak} fixed at $0.1$ and the remaining mass adjusted, increases their total completion probability from $0.5$ to $0.6$. Nevertheless, the interior loss rises from $0.6895$ to $0.8841$ because the allocation further deviates from the teacher's preference. Thus, increasing completion of the representative constraint need not improve matching to the teacher's separate token targets.

\textbf{ESCD} retains the same root projection here and directly supervises native completion-set probabilities at the visited child. Its child objective weights residual completion by teacher source mass without prescribing allocation among valid completions. The construction proceeds as follows.

First, exclude teacher content tokens with an exact single-token student counterpart. This removes \texttt{Cat} from child-event construction while retaining its root supervision. Grouping the remaining candidates by byte-prefix comparability gives
\begin{equation}
\begin{aligned}
V_1&=\{\texttt{Bread},\texttt{Breads}\},
&y_1&=\texttt{Bread},
&M_1&=0.4+0.2=0.6,\\
V_2&=\{\texttt{Break}\},
&y_2&=\texttt{Break},
&M_2&=0.3.
\end{aligned}
\label{eq:worked-example-escd-groups}
\end{equation}
The representative is the shortest member of each group. Although \texttt{Bread} and \texttt{Break} share \texttt{Br}, neither complete string is a prefix of the other, so they remain separate. Sharing a compatible first action is insufficient for aggregation.

Second, check whether the sampled action partially realizes each representative. Since \texttt{Br} is a strict prefix of both $y_1$ and $y_2$, the groups leave different residuals at the same visited child $x'=(x,\texttt{Br})$:
\begin{equation}
r_1=y_1[|\texttt{Br}|:]=\texttt{ead},
\qquad
r_2=y_2[|\texttt{Br}|:]=\texttt{eak}.
\end{equation}
The first residual is defined by the representative \texttt{Bread}; \texttt{eads} is not retained as a separate teacher residual for that group.

Third, collect every student content token whose bytes begin with the entire residual. Assume the complete sets in this toy vocabulary are
\begin{equation}
\begin{aligned}
\mathcal C(r_1)&=\{\texttt{ead},\texttt{eads}\},\\
\mathcal C(r_2)&=\{\texttt{eak}\}.
\end{aligned}
\end{equation}
Both \texttt{ead} and \texttt{eads} complete $r_1$: the former ends at the representative boundary, while the latter extends beyond it. Membership depends on student token bytes, without requiring a one-to-one correspondence with teacher group members.

Using the existing logits at $x'$ gives
\begin{equation}
\begin{aligned}
P_1&=P_\theta(\mathcal C(r_1)\mid x')=0.2+0.3=0.5,\\
P_2&=P_\theta(\mathcal C(r_2)\mid x')=0.1.
\end{aligned}
\end{equation}
These probabilities are not renormalized over either completion set or their union; the remaining probability $0.4$ stays outside both events. Completion candidates require no separate sampling, and the next sampled token need not belong to either set for the child loss to apply.

Finally, weight each completion event's negative log-probability by its root teacher mass:
\begin{equation}
\begin{aligned}
\ell_{\mathrm{child}}
&=-0.6\log\!\left[
p_\theta(\texttt{ead}\mid x')
+p_\theta(\texttt{eads}\mid x')
\right]\\
&\quad-0.3\log p_\theta(\texttt{eak}\mid x')\\
&=-0.6\log0.5-0.3\log0.1
\approx1.1067.
\end{aligned}
\label{eq:worked-example-escd-child}
\end{equation}
The weights remain $0.6$ and $0.3$, without normalization by the compatible teacher mass $0.9$ or division by the student's action probability $0.5$. ESCD combines this child loss with the root projection loss, giving $0.4597+1.1067\approx1.5664$ before common training reduction.

In the probability-halving example used for SimCT, the completion probabilities fall from $(P_1,P_2)=(0.5,0.1)$ to $(0.25,0.05)$. With teacher weights unchanged, the ESCD child loss increases from $1.1067$ to $1.7305$, while the local SimCT loss remains unchanged. Because ESCD uses full-vocabulary probabilities without candidate-set renormalization, the decrease in completion probability directly increases its child loss.

In the redistribution example used for BPM, changing the probabilities of \texttt{ead} and \texttt{eads} from $(0.2,0.3)$ to $(0.1,0.5)$ raises $P_1$ from $0.5$ to $0.6$, with $P_2=0.1$ unchanged. The ESCD child loss decreases from $1.1067$ to $0.9973$, although BPM's loss at $x'$ increases. ESCD therefore assigns a lower loss to improved completion of the representative constraint, even though the allocation moves further from the teacher's token-level preference. Any redistribution preserving $P_1$ and $P_2$ leaves its child loss unchanged.
This flexibility comes from coarsening the supervision: the mass of \texttt{Breads} supports completion of the representative \texttt{Bread}, without separately requiring its final \texttt{s}. The completion set groups actions satisfying the same residual constraint, which need not represent equivalent full continuations. ESCD remains restricted to one additional-token completion, with no child loss when the set is empty.

\subsection{Training Algorithm}
\label{sec:framework-training}

Algorithm~\ref{alg:framework-escd} summarizes ESCD within the standard on-policy distillation pipeline. ESCD reuses the Student rollout, Student training forward pass, and aligned Teacher outputs, and adds completion supervision only at child states already visited by the rollout.
The ESCD-specific computation consists only of sparse event operations on quantities already available from root-level distillation. Teacher events are aggregated by prefix relation, matched against the Student action actually taken, and converted into one-step completion sets at the corresponding visited child state. The resulting loss accesses only the selected Student log-probabilities, avoiding both a dense cross-vocabulary child target and any additional model execution.

  \begin{algorithm}[hbt]
  \caption{Event-Set Completion Distillation on one student response}
  \label{alg:framework-escd}
  \begin{algorithmic}[1]
  \Require Student rollout $A=(a_1,\ldots,a_L)$, aligned teacher hidden states $H_T$, student logits $Z_S$, tokenizer byte artifact $\mathcal A$
  \Require Teacher candidate selector $\textsc{SelectTeacherEvents}$, $\lambda_{\mathrm{ESCD}}=1$
  \Ensure Response-aligned loss vector $\ell_{1:L}$

  \State $\ell_{1:L}\gets 0$; $L_S\gets\operatorname{LogSoftmax}(Z_S)$
  \State $R\gets\textsc{AlignedRootRows}(A,\mathcal A)$

  \Statex \textcolor{gray}{\textbf{[Root distillation]}}
  \ForAll{$r\in R$}
      \State $(v_r,q_r)\gets\textsc{SelectTeacherEvents}(H_T[r])$ \Comment{full vocabulary in our experiments}
      \State $Q_{\mathrm{root}}[r]\gets\textsc{CompileRootTarget}(v_r,q_r,\mathcal A)$
      \State $\ell[r]\gets\textsc{RootDistillationLoss}(Q_{\mathrm{root}}[r],L_S[r])$
  \EndFor

  \Statex \textcolor{blue}{\textbf{[ESCD] Visited-child completion}}
  \State $I_S\gets\textsc{BuildOrLoadBytePrefixIndex}(\mathcal A)$
  \State $(\widetilde v,\widetilde q)\gets\textsc{GatherCandidateEventsAcrossCPRanks}(v,q)$
  \ForAll{$r\in R$ with a valid next prediction row in the same response}
      \State $a\gets A[r]$; $c\gets r+1$ \Comment{logical response coordinates}
      \State $\mathcal G_r\gets\textsc{PrefixQuotient}\bigl(\textsc{EligibleNonSharedEvents}(\widetilde v[r],\widetilde q[r],\mathcal A)\bigr)$
      \ForAll{$(y_g,M_g)\in\mathcal G_r$ such that $b_S(a)\prec y_g$}
          \State $C_g\gets I_S.\textsc{CompletionTokens}\bigl(y_g[|b_S(a)|:]\bigr)$
          \If{$C_g\neq\varnothing$}
              \State $\ell[c]\gets\ell[c]-\lambda_{\mathrm{ESCD}}M_g\operatorname{LogSumExp}_{u\in C_g}L_S[c,u]$
          \EndIf
      \EndFor
  \EndFor
  \State \Return $\ell_{1:L}$
  \end{algorithmic}
  \end{algorithm}

\clearpage
\section{Additional Analysis of ESCD}
\label{app:escd-analysis}

Section~\ref{sec:escd-ablation} reports two main observations: preserving the completion set yields higher local update fidelity than selecting a single student token, and child supervision is frequently available along student trajectories. This appendix analyzes conditioning states, event structures, boundary effects, and completion depth, alongside reasoning patterns, OPD degradation, and SFT-associated changes.

Specifically, we examine the following questions:

\begin{enumerate}
    \item \textbf{Can the effect of the conditioning state be isolated?} (\ref{app:conditioning-state}) We examine the available support for comparing naturally visited and unvisited compatible child states while controlling event identity and completion depth.

    \item \textbf{Which event structures dominate in practice?} (\ref{app:event-geometry}) We classify events by teacher group size and the number of compatible student actions at the root, comparing their teacher-mass shares and conditional one-step completion coverage across tokenizer pairs.

    \item \textbf{How much do boundary-crossing completions affect supervision?} (\ref{app:boundary-crossing}) We measure the probability assigned to tokens extending beyond the residual event boundary and examine how removing them changes the local update.

    \item \textbf{How often is child supervision available, and is one step enough?}(\ref{app:coverage-depth}) We quantify the teacher probability mass reached by compatible on-policy actions and the fraction requiring more than one additional student token to complete.

    \item \textbf{How do rollout-level reasoning patterns vary across models?} (\ref{app:rollout-geometry}) We use PCA to describe the overlap and variation of response-level reasoning features across models within each benchmark.

    \item \textbf{How does OPD degradation manifest under large Teacher--Student scale disparity?} (\ref{app:opd-degeneration}) We document repetitive continuations and premature termination through rollout statistics and saved responses, showing that termination alone does not imply task completion.

    \item \textbf{What changes in reasoning performance and behavior accompany SFT?} (\ref{app:sft-stabilization}) We compare aggregate scores and matched-prompt responses from the teacher, base student, and SFT student, focusing on derivation structure and explicit consistency checks .
\end{enumerate}

\subsection{Does the Choice of Child State Matter?}
\label{app:conditioning-state}

When a student action partially realizes a teacher event, ESCD supervises the residual at the resulting child state. Other compatible actions can lead to different child states and residuals. We ask whether the visited state provides better completion supervision than these alternatives.

\paragraph{Comparison setup.}

We compare the child reached by the sampled action with a child that would have been reached under another compatible action. We call the latter a counterfactual child because the rollout did not actually take that action. Both branches are evaluated for completion of the same teacher event, with root supervision held fixed. The alternative action is selected without using COUF scores or downstream results.
To isolate the effect of the child state, both branches should also use the same completion depth. For the one-step objective studied here, this requires both residuals to admit completion by one additional student token. Otherwise, a difference in supervision could reflect the number of continuation steps rather than the choice of state.

\paragraph{Available comparisons.}

The current data provide few examples meeting these requirements. We identify 32 Qwen and 37 GLM rollout contexts where the visited child state and a counterfactual child state can be matched for the same teacher event. These pairs cover little teacher probability mass, and none allows both branches to complete the same event in one additional step (Table~\ref{tab:conditioning-state-support}). Thus, the available data do not provide a comparison that simultaneously controls event identity and one-step completion depth.

\begin{table}[hbt]
\centering
\small
\setlength{\tabcolsep}{6pt}
\caption{Available comparisons between visited and alternative child states. The last column counts contexts where both branches admit one-step completion of the same teacher event.}
\label{tab:conditioning-state-support}
\begin{tabular}{lccc}
\toprule
\textbf{Teacher $\rightarrow$ Student}
& \textbf{Contexts}
& \textbf{Matched mass}
& \textbf{Both 1-step} \\
\midrule
Qwen3 $\rightarrow$ Qwen3.5
& 32 & 0.8745\% & 0 \\
GLM-Z1 $\rightarrow$ Qwen3.5
& 37 & 0.0662\% & 0 \\
\bottomrule
\end{tabular}
\end{table}

\paragraph{Comparison allowing deeper completion.}

To evaluate the available alternatives, we allow the counterfactual branch to continue for additional steps when one-step completion is unavailable. The visited branch retains ESCD's one-step objective. Table~\ref{tab:conditioning-state-couf} reports agreement with the reference logit gradient using the COUF metric defined in Section~\ref{sec:escd-ablation}. Under this relaxed comparison, the counterfactual branch achieves higher COUF than the visited branch in both model pairs.

\begin{table}[t]
\centering
\small
\caption{State-space COUF when the counterfactual branch may use deeper completion. The teacher event and root supervision are matched, but completion depth is not controlled.}
\label{tab:conditioning-state-couf}
\begin{tabular}{lccc}
\toprule
\textbf{Teacher $\rightarrow$ Student}
& \textbf{Root}
& \textbf{Counterfactual}
& \textbf{Visited} \\
\midrule
Qwen3 $\rightarrow$ Qwen3.5
& 0.999031
& 0.999982
& 0.999032 \\
GLM-Z1 $\rightarrow$ Qwen3.5
& 0.998486
& 0.999997
& 0.998489 \\
\bottomrule
\end{tabular}
\end{table}

These results do not establish which child state is preferable: the branches differ in both conditioning state and allowed completion depth, and the comparison covers little teacher mass. ESCD uses visited states because their logits are already available from the student training forward pass, allowing completion supervision without additional rollouts. The present analysis does not establish an independent supervision advantage of these states over other compatible states.

\subsection{What Event Structures Does ESCD Encounter?}
\label{app:event-geometry}

An event may admit several compatible student actions at the root. Some produce only a prefix of the event, while others complete it immediately. When the sampled action produces a strict prefix, the student reaches a child state with a residual byte constraint. Different prefix actions can leave different residuals, so compatibility at the root does not guarantee that the event can be completed in one additional step. We therefore distinguish the number of compatible first actions from the number of valid completions after the sampled action. The analysis below characterizes the former and measures how often the latter is nonzero along student trajectories.

\paragraph{Event structure at the root.}

For a teacher group $g$, let $m_g=|V_g|$ denote its size and $y_g$ its representative byte string. We partition compatible student first actions into those that partially realize $y_g$ and those that complete it:
\begin{equation}
\begin{aligned}
\mathcal A(y_g)
&=\{u\in V_S^{\mathrm{cont}}:b_S(u)\prec y_g\},\\
\mathcal T(y_g)
&=\{u\in V_S^{\mathrm{cont}}:y_g\preceq b_S(u)\},\\
b_g
&=|\mathcal A(y_g)|+|\mathcal T(y_g)|.
\end{aligned}
\label{eq:root-event-compatibility}
\end{equation}
Actions in $\mathcal A(y_g)$ leave a nonempty residual, whereas those in $\mathcal T(y_g)$ complete the event exactly or extend beyond its boundary. Their combined count $b_g$ describes the choices available at the root.

After sampling $a\in\mathcal A(y_g)$, $n_g(a)=|\mathcal C(r_g)|$ counts valid one-step completions at the child state (Appendix~\ref{sec:framework-escd}). The counts describe different stages: a single compatible first action ($b_g=1$) may leave a residual with zero, one, or multiple completions.

\paragraph{Coverage along student trajectories.}

We analyze the full teacher vocabulary at aligned, non-whitespace positions in 16 frozen student trajectories per pair. After removing invalid tokens and exact byte matches to student content tokens, we aggregate candidates by prefix compatibility. Table~\ref{tab:geometry-full-trajectory} groups event occurrences by $m_g$ and $b_g$, restricting attention to $\mathcal A(y_g)\neq\varnothing$. Each occurrence is counted at its rollout position and weighted by teacher group mass $M_g$.

\emph{Mass share} gives each class's proportion of analyzed teacher mass. \emph{Child visited} is the within-class mass fraction with a sampled action in $\mathcal A(y_g)$. \emph{One-step given visit} is the fraction of visited mass admitting at least one valid one-step completion, measuring availability rather than sampled success. Overall values use pooled masses across classes.

\begin{table}[t]
\centering
\small
\setlength{\tabcolsep}{4pt}
\renewcommand{\arraystretch}{1.1}
\caption{Root event structure and completion availability from the full teacher vocabulary after exact-shared and invalid-token filtering. Statistics use aligned, non-whitespace positions in 16 frozen student trajectories per pair, grouped by $m_g$ and $b_g$.}
\vspace{-3mm}
\label{tab:geometry-full-trajectory}

\begin{tabular}{@{}lccccc@{}}
\toprule
\textbf{Pair}
& $\boldsymbol{m_g}$
& $\boldsymbol{b_g}$
& \textbf{Mass share}
& \textbf{Child visited}
& \textbf{One-step given visit} \\
\midrule
\multirow{5}{*}{Qwen3 $\rightarrow$ Qwen3.5}
& $1$  & $1$
& 0.0026\% & 34.5351\% & 96.0012\% \\

& $1$  & $>1$
& 0.1089\% & 70.6500\% & 71.9885\% \\

& $>1$ & $1$
& \textbf{88.2125\%} & 80.9535\% & 99.9991\% \\

& $>1$ & $>1$
& 11.6760\% & 88.6045\% & 92.4118\% \\

& \multicolumn{2}{c}{\textit{Overall}}
& 100.0000\% & \textbf{81.8344\%}
& \textbf{99.0135\%} \\

\midrule

\multirow{5}{*}{GLM-Z1 $\rightarrow$ Qwen3.5}
& $1$  & $1$
& 2.3143\% & 88.8103\% & 99.9993\% \\

& $1$  & $>1$
& 0.0917\% & 51.3026\% & 80.4281\% \\

& $>1$ & $1$
& \textbf{90.1026\%} & 82.6607\% & 100.0000\% \\

& $>1$ & $>1$
& 7.4914\% & 89.5888\% & 93.0875\% \\

& \multicolumn{2}{c}{\textit{Overall}}
& 100.0000\% & \textbf{83.2933\%}
& \textbf{99.4319\%} \\

\bottomrule
\end{tabular}
\end{table}

Groups with multiple teacher candidates and one compatible student first token ($m_g>1$, $b_g=1$) account for 88.21\% of analyzed mass for Qwen and 90.10\% for GLM. Nearly all their visited mass admits one-step completion. Lower-coverage classes contribute much less visited mass, yielding overall conditional coverage of 99.01\% and 99.43\%, respectively. This describes supervision availability, without attributing downstream gains to individual classes.

\subsection{Does the Completion Set Become Too Coarse at the Event Boundary?}
\label{app:boundary-crossing}

The event-set construction in ESCD deliberately avoids choosing a single student realization. This raises a potential ambiguity when a student token completes the residual teacher event but also extends beyond its boundary. For example, if the residual bytes correspond to \texttt{Memory}, both \texttt{Memory} and \texttt{MemoryWarning} satisfy the current prefix-compatible completion rule.
Importantly, this does not mean that the two tokens are treated as semantically equivalent. ESCD supervises the byte-prefix event that the continuation completes the residual, rather than requiring the next student token to terminate exactly at the same boundary. A longer token can therefore satisfy the event while additionally committing bytes that are not specified by the current teacher event.

\paragraph{How much probability crosses the boundary?}

To quantify whether this ambiguity is substantial in practice, we separate exact completions from tokens that strictly extend beyond the residual. We report the fraction of the full completion-event probability assigned to the latter:

\begin{equation}
\rho_{\mathrm{cross}}
=
\frac{
\sum_{u:\,r\prec b_S(u)}
p_\theta(u\mid x,a)
}{
\sum_{u:\,r\preceq b_S(u)}
p_\theta(u\mid x,a)
}.
\label{eq:boundary-crossing-ratio}
\end{equation}

A large completion set does not necessarily imply a large $\rho_{\mathrm{cross}}$: many prefix-compatible tokens may exist in the vocabulary while receiving negligible student probability. We therefore evaluate $\rho_{\mathrm{cross}}$ on naturally visited strict-prefix events whose child logits are available for computation, resulting in 48 Qwen and 287 GLM events.

\begin{table}[t]
\centering
\small
\setlength{\tabcolsep}{6pt}
\caption{Boundary-crossing ambiguity on materialized natural-child events. ``Set size'' is weighted by teacher source mass; ``Crossing prob.'' is the fraction of completion-event probability assigned to tokens that extend beyond the residual boundary.}
\vspace{-3mm}

\label{tab:boundary-crossing-main}
\begin{tabular}{lcccc}
\toprule
\textbf{teacher $\rightarrow$ Student}
& \textbf{Events}
& \textbf{Set size}
& \textbf{Full prob.}
& \textbf{Crossing prob.} \\
\midrule
Qwen3 $\rightarrow$ Qwen3.5
& 48
& 35.28
& 0.8846
& \textbf{2.10\%} \\

GLM-Z1 $\rightarrow$ Qwen3.5
& 287
& 14.05
& 0.9412
& \textbf{0.63\%} \\
\bottomrule
\end{tabular}
\end{table}

Table~\ref{tab:boundary-crossing-main} reveals an important distinction between tokenizer topology and realized student probability. Qwen has an average completion-set size above 35, yet boundary-crossing tokens carry only 2.10\% of the completion probability; for GLM, the corresponding fraction is only 0.63\%. Thus, a large prefix-compatible set does not imply that probability is broadly distributed across its members.

For Qwen, this large set size is primarily driven by very short residuals: 45 of the 48 materialized events contain a one-byte residual, with newline being the dominant case. Short residuals naturally induce large prefix cylinders, but the student probability remains strongly concentrated near the exact completion.

\paragraph{What does the event loss leave unidentified?}

Although the realized crossing probability is small, the full event-set objective has a genuine identifiability limitation. The child loss depends only on the total probability assigned to the completion set:

\begin{equation}
\mathcal L_{\mathrm{child}}
=
-M
\log
P_\theta(\mathcal C_{\mathrm{full}}(r)\mid x,a).
\label{eq:boundary-crossing-null}
\end{equation}

Consequently, redistributing probability among tokens inside the same completion set while keeping their total probability fixed leaves this loss unchanged. For example, assigning completion probability $(0.8,0.1)$ to an exact and a crossing token produces the same child loss as $(0.1,0.8)$ whenever the total event probability is unchanged.
This is a property of event marginalization rather than evidence of an observed failure. It means that ESCD identifies the probability of completing the teacher byte event, but does not locally identify how this probability should be allocated among different student realizations that already satisfy the event.

\paragraph{Does removing crossing completions materially change the local update?}

We next compare the current \textbf{Full} target with a stricter \textbf{Exact} target that retains only student tokens ending exactly at the residual boundary. We also consider a \textbf{Minimal} target that keeps the shortest prefix-compatible completions.
In the current materialized subsets, every evaluated event has an exact one-step completion, and the shortest compatible completion is always exact. Therefore, \textbf{Minimal and Exact are identical on all evaluated events}; the present diagnostic can distinguish only Exact/Minimal from Full.
We compute the local logit gradient induced by each target on the same student child logits and compare the Exact/Minimal gradient with the Full gradient.

\begin{table}[t]
\centering
\small
\setlength{\tabcolsep}{6pt}
\caption{Local target-gradient sensitivity to removing boundary-crossing completions.}
\vspace{-3mm}

\label{tab:boundary-crossing-gradient}
\begin{tabular}{lccc}
\toprule
\textbf{Teacher $\rightarrow$ Student}
& \textbf{Projection}
& \textbf{Energy ratio}
& \textbf{Cosine} \\
\midrule
Qwen3 $\rightarrow$ Qwen3.5
& 1.0566
& 1.1550
& \textbf{0.9832} \\

GLM-Z1 $\rightarrow$ Qwen3.5
& 1.0302
& 1.0857
& \textbf{0.9887} \\
\bottomrule
\end{tabular}
\end{table}

Removing boundary-crossing tokens increases the local gradient magnitude slightly, but the resulting direction remains highly aligned with the Full target: the gradient cosine is 0.9832 for Qwen and 0.9887 for GLM. In the current frozen natural-child subsets, boundary crossing therefore changes the local target quantitatively but does not induce a substantially different update direction.

\paragraph{Takeaway.}

The analysis leads to three conclusions. First, boundary-crossing tokens are consistent with ESCD's byte-prefix event semantics and should not be interpreted as an implementation error or as semantically equivalent tokens. Second, the full event-set objective does leave probability allocation within the completion set under-specified, but the realized crossing probability is small in the current Qwen and GLM subsets. Third, removing crossing tokens changes gradient magnitude but preserves a highly similar local update direction.
We therefore view boundary crossing as a genuine source of local target coarsening, but not as an observed failure mode in the current frozen analysis. Determining whether stricter boundary-aware targets improve training would require native Exact/Minimal/Full runs under a common training protocol; in particular, events without an exact one-step completion are needed to distinguish Minimal from Exact.

\subsection{How Often Is Child Supervision Available, and Is One Step Enough?}
\label{app:coverage-depth}

We further examine the availability of child supervision and the scope of one-step completion. The analysis uses the full teacher vocabulary at aligned, non-whitespace positions in 16 frozen student trajectories per pair: 374809 positions for Qwen and 365153 for GLM. All statistics follow the candidate filtering and prefix aggregation described in the main text. We pool teacher-mass numerators and denominators across positions and trajectories before computing percentages.

A representative event \(y_g\) is branchable if some student action produces a nonempty strict prefix of its bytes. At each position, a continuing student action is scored by the total teacher mass of compatible branchable representative events. The three visitation statistics in Table~\ref{tab:full-coverage-funnel} share a denominator: total branchable teacher mass across all analyzed positions.

\textbf{Position upper bound} counts all branchable mass at a position whenever the sampled action matches any continuing branch. \textbf{Observed} counts only the event mass compatible with the sampled action. \textbf{Top-1 observed} counts the highest-scoring branch's mass only when that branch is sampled. Branches are ranked by teacher source mass, not student action probability, with ties broken by ascending student token ID. These are mass fractions, not position hit rates; the position upper bound can include events incompatible with the sampled action.

\begin{table}[hbt]
\centering
\small
\caption{Compatible-child visitation under full-vocabulary analysis. All columns are relative to the same pooled branchable teacher mass. The position upper bound counts all branchable mass at positions with any compatible sampled action; Top-1 ranks branches by teacher source mass.}
\label{tab:full-coverage-funnel}
\begin{tabular}{lccc}
\toprule
\textbf{Teacher $\rightarrow$ Student}
& \textbf{Position upper bound}
& \textbf{Observed}
& \textbf{Top-1 observed} \\
\midrule
Qwen3 $\rightarrow$ Qwen3.5
& 93.47\%
& \textbf{81.83\%}
& 71.55\% \\
GLM-Z1 $\rightarrow$ Qwen3.5
& 95.49\%
& \textbf{83.29\%}
& 76.23\% \\
\bottomrule
\end{tabular}
\end{table}

Compatible children are naturally visited for 81.83\% and 83.29\% of branchable teacher mass for Qwen and GLM, respectively. Thus, over 80\% of branchable mass is compatible with sampled student actions. This measures visitation, not completion by the next sampled token. Visitation does not guarantee that the residual admits one-step completion.

We measure completion availability conditional on this visited teacher mass. For each visited representative event, the sampled action leaves a residual \(r_g\). \textbf{1-step} is the mass fraction for which the completion set \(\mathcal C(r_g)\) is nonempty, and \textbf{Deeper} is the remaining fraction. These quantities measure whether a next-token completion exists in the student vocabulary, rather than how often the student samples one. The completion criterion applies to the representative residual after prefix aggregation.

\begin{figure*}[t]
    \centering
    \includegraphics[width=\textwidth]{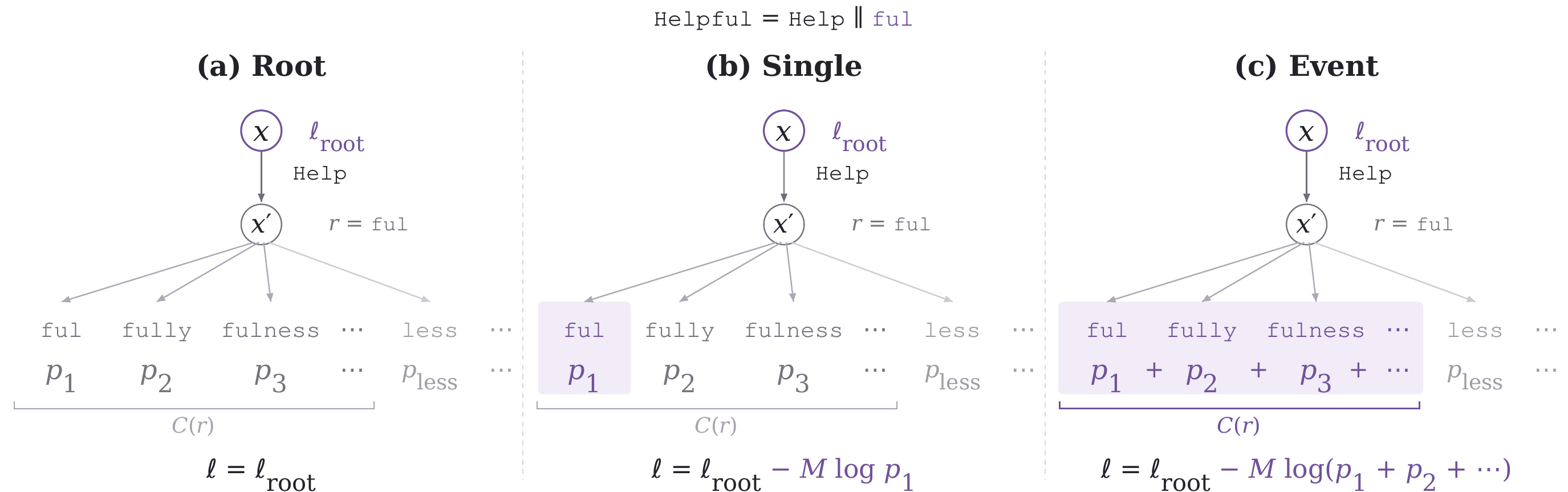}
    \caption{\textbf{One-step completion and supervision choices.} Sampling \texttt{Help} leaves residual $r=\texttt{ful}$ for the representative event \texttt{Helpful}. Valid one-token completions include \texttt{ful}, \texttt{fully}, and \texttt{fulness}; \texttt{less} lies outside $\mathcal C(r)$. All three objectives share the same root loss and student predictions. \textbf{(a) Root} adds no child loss. \textbf{(b) Single} supervises one selected completion. \textbf{(c) Event} supervises the completion set's total probability. Here, $p_i$ denotes a candidate's native student probability at $x'$, and $M$ is the representative group's teacher mass. Ellipses indicate additional candidates. Tokenizations and the Single selection are illustrative; losses are shown for one group before training-mask reduction.}
    \label{fig:root-single-event-appendix}
\end{figure*}

One-step completion may admit multiple valid tokens. Figure~\ref{fig:root-single-event-appendix} illustrates how Root, Single, and Event use the same visited state: Root adds no child loss, Single supervises one selected valid completion, and Event supervises the total probability of the completion set. Single and Event therefore differ in their supervision targets, not in completion depth.

\begin{table}[t]
\centering
\small
\caption{One-step completion availability for representative residual constraints, weighted by teacher mass and conditional on a compatible child being visited.}
\label{tab:completion-depth}
\begin{tabular}{lcc}
\toprule
\textbf{Teacher $\rightarrow$ Student}
& \textbf{1-step}
& \textbf{Deeper} \\
\midrule
Qwen3 $\rightarrow$ Qwen3.5
& \textbf{99.01\%}
& 0.99\% \\
GLM-Z1 $\rightarrow$ Qwen3.5
& \textbf{99.43\%}
& 0.57\% \\
\bottomrule
\end{tabular}
\end{table}

Table~\ref{tab:completion-depth} shows that representative residuals covering 99.01\% of visited Qwen mass and 99.43\% of visited GLM mass admit completion by one additional student token. The remaining 0.99\% and 0.57\% lack one-step completion and contribute no auxiliary child loss under ESCD.

These measurements support the availability of ESCD's local child supervision in the studied tokenizer pairs and trajectories. Coverage depends on the filtering, aggregation, and visitation criteria above; it does not establish one-step completion of every original teacher token or satisfaction of longer group members' full byte requirements. Residuals without one-step completion remain a small but nonzero share of observed mass, motivating deeper completion.

\subsection{Rollout-Level Reasoning Geometry}
\label{app:rollout-geometry}
\begin{figure*}[!hbt]
    \centering
    \includegraphics[width=\textwidth]{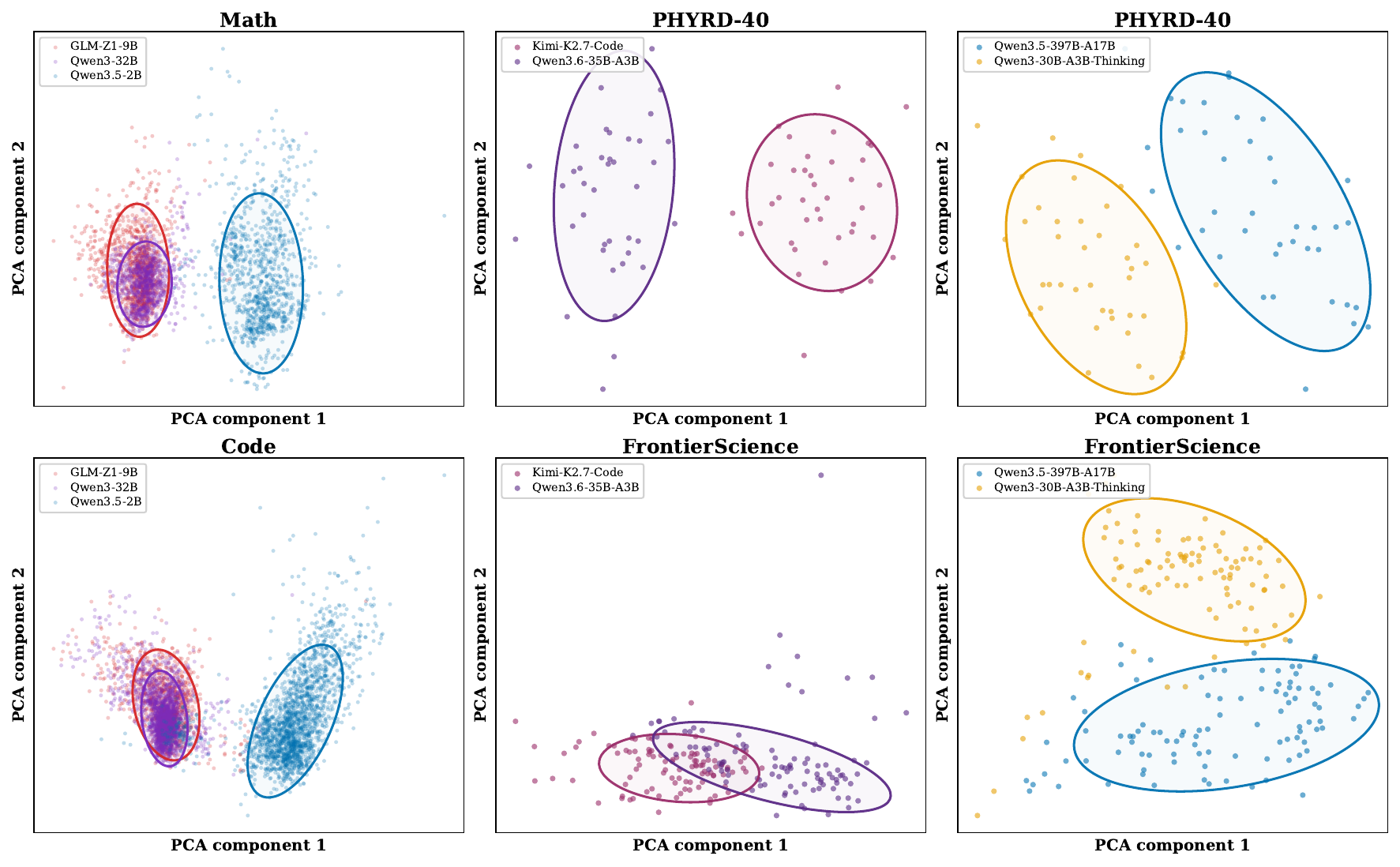}
    \vspace{-6mm}
\caption{PCA projections of response-level reasoning features across math, code, and scientific reasoning benchmarks. Each point represents a generated response, with colors identifying models; ellipses summarize response distributions after robust trimming. PCA is fitted separately in each panel, so distances and directions are comparable only within panels.}
    \label{fig:rollout-reasoning-pca}
\end{figure*}

Before examining individual training regimes, we use PCA to visualize variation in response-level reasoning signatures across models and benchmarks (Figure~\ref{fig:rollout-reasoning-pca}). The projections cover Math, Code, PHYRD-40, and FrontierScience, providing a qualitative overview of the overlap and variation among teacher and student responses within each benchmark and model group.

\subsection{OPD Degradation under Large Teacher--Student Scale Disparity}
\label{app:opd-degeneration}

We observe severe generation degradation during direct BPM-based OPD from Kimi-K2.7-Code to Qwen3.6-35B-A3B, a pairing with a large disparity in total parameter count. Without additional SFT initialization, the student exhibits two failure modes: repetitive continuations that exhaust the decoding budget and premature termination before substantive reasoning. We document these failures using training statistics and saved responses.

\paragraph{Repetitive continuation and premature termination.}

At rollout steps 1--19, no responses are length-truncated or flagged by the native repetition detector. At step 20, one of 16 responses is both repetition-flagged and length-truncated; at step 21, two of 16 responses meet both criteria. All three reach the 80,000-token limit, with repetitive passages consisting of the symbol $\mathbb{Z}$, the phrase \emph{``classifies by Weyl-Equivalence''}, or malformed LaTeX fragments. Premature termination also occurs at step 20: one response ends after only 12 tokens, briefly restating the topic without beginning substantive reasoning. These observations show that excessive repetition and premature termination can coexist at the same training stage.

\paragraph{Response-length collapse and premature termination.}

In the truncation-filtered run, retained responses become markedly shorter during training (Table~\ref{tab:opd-degeneration}), with mean length falling from 10,587 tokens at step 16 to 8 at step 32. All 16 responses at step 32 contain at most 20 tokens, consisting of unfinished preambles, task fragments, or special tokens followed by termination. Outputs remain short at step 38. The saved text confirms that this shortening reflects termination before substantive reasoning rather than more concise solutions.

\begin{table}[hbt]
\centering
\small
\setlength{\tabcolsep}{8pt}
\caption{Response-length collapse in the truncation-filtered Kimi-to-Qwen BPM run. Statistics describe the 16 responses retained at each listed step, rounded to the nearest token.}
\vspace{-3mm}
\label{tab:opd-degeneration}
\begin{tabular}{rrrr}
\toprule
\textbf{Step}
& \textbf{Mean tokens}
& \textbf{Median tokens}
& \textbf{Max tokens} \\
\midrule
1  & 7,698  & 7,550 & 14,470 \\
16 & 10,587 & 7,605 & 50,590 \\
20 & 3,022  & 2,518 & 8,099  \\
26 & 247    & 24    & 1,818  \\
30 & 56     & 8     & 566    \\
32 & 8      & 6     & 20     \\
38 & 20     & 19    & 52     \\
\bottomrule
\end{tabular}
\end{table}

\paragraph{Representative responses.}

Together, these cases reveal a stability challenge for direct OPD in this Teacher--student pairing: responses may exhaust the decoding budget through repetition or terminate before substantive reasoning. These failures motivate examining SFT as a preparatory stage for subsequent OPD. We next analyze the SFT checkpoint's reasoning performance and response structure to characterize the initialization used in our large-scale distillation experiments.

\begin{tcolorbox}[
    trainingcase,
    colbacktitle=teal!11,
    title={Repetitive continuation and premature termination}
]

\textcolor{glmviolet}{\textbf{Repetitive continuation}}

\vspace{1mm}

{\color{textgray}\normalfont
\textbf{Unfiltered run, step 20.}
The prompt asks for an analysis connecting linear angle evolution to a Fourier representation of late-time density. The response initially discusses this connection but later repeatedly emits:
\[
\mathbb{Z}\mathbb{Z}\mathbb{Z}\mathbb{Z}
\mathbb{Z}\mathbb{Z}\mathbb{Z}\mathbb{Z}\cdots
\]
\textit{The repeated passage is abbreviated. The response reaches the 80,000-token limit and is recorded as length-truncated.}
}

\vspace{1.5mm}
\hrule
\vspace{1.5mm}

\textcolor{glmviolet}{\textbf{Premature termination}}

\vspace{1mm}

{\color{textgray}\normalfont
\textbf{Filtered run, step 32.}
Three saved responses to distinct mathematical-physics prompts are shown below:

\vspace{1mm}

\textbf{Example 1:}
\texttt{We need<|im\_end|>}

\vspace{1mm}

\textbf{Example 2:}
\texttt{Here<|im\_end|>}

\vspace{1mm}

\textbf{Example 3:}
\texttt{</think><|im\_end|>}

\vspace{1.5mm}

\textit{All 16 retained responses at this step contain at most 20 tokens and fail to provide substantive solutions. No length-truncated responses are discarded.}
}

\end{tcolorbox}

\subsection{Reasoning Performance and Structure after SFT}
\label{app:sft-stabilization}

We examine the Qwen3.6-35B-A3B checkpoint after supervised fine-tuning, which improves aggregate long-form reasoning performance over the base student. Matched-prompt examples also show a recurring organizational difference: normalization choices, boundary conditions, limiting cases, and local consistency checks are more often explicit before the derivation concludes.
For SFT initialization, we use Kimi-K2.7-Code to generate approximately 1,800 trajectories from the OPD training prompts and fine-tune the student on these trajectories for three epochs.

\paragraph{Evidence and scope.}
On PHYRD-40, the Qwen3.6-35B-A3B base model scores 59.2, the SFT checkpoint 66.2, and Kimi-K2.7-Code 83.6. Their FrontierScience Olympiad accuracies are 61.0\%, 68.0\%, and 75.0\%, respectively. SFT improves the student's mean rubric score by 7.0 points and Olympiad accuracy by 7.0 percentage points, yielding a more capable starting checkpoint for subsequent OPD. Matched-prompt examples complement these results by comparing how the teacher, base student, and SFT student handle assumptions, intermediate derivations, and consistency checks, characterizing the SFT student's reasoning behavior before OPD.

\subsubsection{Matched-Prompt Reasoning Patterns}
\label{app:reasoning-pattern-cases}

We compare lightly trimmed final-answer fragments from three shared PHYRD-40 prompts. The excerpts are drawn only from the models' visible responses, and all selected generations terminate normally. Three examples are not sufficient for a model ranking; they are used here to inspect how the teacher, base student, and SFT student structure long derivations, represent intermediate quantities, perform local checks, and close the requested argument.

\paragraph{A033: Linear displacement kernel.}

The task requires the progression \(\Delta\mathbf s\rightarrow C_{ij}(q)\rightarrow\{A(q),B(q)\}\rightarrow K(\mathbf q,\mathbf k)\), together with consistency checks such as \(q\to0\) and \(\mathbf k=0\).

\begin{tcolorbox}[
    trainingcase,
    colbacktitle=glmviolet!12,
    title={Kimi-K2.7-Code}
]
\textcolor{glmviolet}{\textbf{Final-answer fragment}}

\vspace{1.5mm}

{\color{textgray}\normalfont
One finds
\[
C_{ij}(\mathbf q)
=
\frac{1}{\pi^2}\int_0^\infty\!dp\,P(p)
\left[
\frac{1-j_0(pq)-j_2(pq)}{3}\,\delta_{ij}
+j_2(pq)\,\hat q_i\hat q_j
\right].
\]
Thus
\[
A(q)
=
\frac{1}{\pi^2}\int_0^\infty\!dp\,P(p)
\frac{1-j_0(pq)-j_2(pq)}{3},
\qquad
B(q)
=
\frac{1}{\pi^2}\int_0^\infty\!dp\,P(p)\,j_2(pq).
\]
At zero separation, \(j_0(0)=1\) and \(j_2(0)=0\), so \(A(0)=B(0)=0\). Substitution gives
\[
K(\mathbf k,\mathbf q)
=
\exp\!\left[
-\frac12
\left(
A(q)k^2+B(q)(\mathbf k\!\cdot\!\hat{\mathbf q})^2
\right)
\right].
\]
}
\end{tcolorbox}

\begin{tcolorbox}[
    trainingcase,
    colbacktitle=red!10,
    title={Qwen3.6-35B-A3B Base}
]
\textcolor{red!65!black}{\textbf{Final-answer fragment}}

\vspace{1.5mm}

{\color{textgray}\normalfont
The linear displacement is a zero-mean Gaussian random field. Therefore
\[
W(\mathbf q,\mathbf k)
=
\exp\!\left[
-\frac12 k_i k_j
\langle
\Delta s_i^{(1)}
\Delta s_j^{(1)}
\rangle
\right].
\]
Decompose
\[
C_{ij}(q)
=
\alpha(q)\delta_{ij}
+
\beta(q)\hat q_i\hat q_j.
\]
Defining transverse and longitudinal contractions \(T(q)\) and \(L(q)\) gives
\[
3\alpha+\beta=2T,
\qquad
\alpha+\beta=2L,
\]
and hence
\[
\alpha=T-L,
\qquad
\beta=3L-T.
\]
Substitution yields the final rotationally invariant kernel.
}
\end{tcolorbox}

\begin{tcolorbox}[
    trainingcase,
    colbacktitle=qwenblue!20,
    title={Qwen3.6-35B-A3B + SFT}
]
\textcolor{qwenblue}{\textbf{Final-answer fragment}}

\vspace{1.5mm}

{\color{textgray}\normalfont
Carrying out the angular integrals gives explicit radial forms for \(A(q)\) and \(B(q)\). The kernel is
\[
K(\mathbf q,\mathbf k;t)
=
\exp\!\left[
-\frac12
\left(
A(q)k^2+B(q)(\mathbf k\!\cdot\!\hat{\mathbf q})^2
\right)
\right].
\]
For \(\mathbf k=0\), the exponent vanishes and \(K=1\). As \(q\to0\), \(\Delta\mathbf s\to0\), so \(A(q),B(q)\sim q^2\) and \(K\to1\). The covariance is positive semidefinite because it is constructed from the squared displacement difference.
}
\end{tcolorbox}

\textbf{Observed pattern.} All three responses reach the same kernel structure. The clearest difference appears after the main derivation: the SFT response explicitly checks \(\mathbf k=0\), the \(q\to0\) limit, and positive semidefiniteness, whereas the base response stops after the compact \(T/L\) construction. Kimi also checks the zero-separation limit, but organizes the derivation directly around the covariance functions.

\paragraph{A065: Coherent-state MPO exponential.}

The task requires iterating a normalized Gaussian source identity into a coherent-state MPO while preserving signs, auxiliary contractions, operator support, and boundary conditions.

\begin{tcolorbox}[
    trainingcase,
    colbacktitle=glmviolet!12,
    title={Kimi-K2.7-Code}
]
\textcolor{glmviolet}{\textbf{Final-answer fragment}}

\vspace{1.5mm}

{\color{textgray}\normalfont
The local tensor is
\[
W_i(\phi_{i-1},\bar\phi_i)
=
\exp\!\left[
\phi_{i-1}^{\alpha}\hat A_i^{\alpha\beta}\bar\phi_i^{\beta}
+\phi_{i-1}^{\alpha}\hat B_i^{\alpha}
+\hat C_i^{\beta}\bar\phi_i^{\beta}
+\hat D_i
\right].
\]
If interactions vanish, the Gaussian integrals decouple and recover the on-site exponential. For two sites, the remaining integral produces \(e^{h_{L_1}\cdot h_{R_1}}\). For a three-site term \(O_1O_2O_3\), choosing \(\hat C_1=O_1\), \(\hat A_2=O_2\), and \(\hat B_3=O_3\) reproduces the expected exponent after two Gaussian contractions. Thus \(\hat A\) propagates the auxiliary channel responsible for longer-range operators.
}
\end{tcolorbox}

\begin{tcolorbox}[
    trainingcase,
    colbacktitle=red!10,
    title={Qwen3.6-35B-A3B Base}
]
\textcolor{red!65!black}{\textbf{Final-answer fragment}}

\vspace{1.5mm}

{\color{textgray}\normalfont
Collecting the fields at one cut gives
\[
-\bar\phi_i\!\cdot\!\phi_i
+h_{L_i}\!\cdot\!\bar\phi_i
+\phi_i\!\cdot\!
\left(
\hat B_{i+1}+\hat A_{i+1}h_{R_{i+1}}
\right).
\]
Applying the source identity yields
\[
\mathcal W_{i+1}(u,v)
=
\exp\!\left[
u\!\cdot\!\hat B_{i+1}
+
u\!\cdot\!\hat A_{i+1}\!\cdot\!v
\right]
\exp\!\left[
\hat C_{i+1}\!\cdot\!v+\hat D_{i+1}
\right].
\]
Iterating over all cuts contracts neighboring auxiliary states and yields the full MPO exponential.
}
\end{tcolorbox}

\begin{tcolorbox}[
    trainingcase,
    colbacktitle=qwenblue!20,
    title={Qwen3.6-35B-A3B + SFT}
]
\textcolor{qwenblue}{\textbf{Final-answer fragment}}

\vspace{1.5mm}

{\color{textgray}\normalfont
The full representation is
\[
e^H
=
e^{H_{L_1}}
\int
\prod_{i=1}^{N-1}
\mathcal D[\phi_i,\bar\phi_i]
\prod_{i=1}^{N-1}
\mathcal W_i
(\phi_i,\bar\phi_i;\phi_{i+1},\bar\phi_{i+1})
e^{H_{R_N}}.
\]
The Gaussian normalization is recovered when all sources vanish. Every quadratic term has sign \(-\bar\phi_i\cdot\phi_i\), while the source identity produces \(+J\cdot\bar J\). For \(N=2\), the remaining Gaussian integral reproduces \(e^{h_{L_1}\cdot h_{R_1}}\), recovering the original cut decomposition.
}
\end{tcolorbox}

\textbf{Observed pattern.} The base response proceeds directly from the cut identity to the iterated representation. The SFT response adds explicit checks of the Gaussian normalization, the sign convention, and the \(N=2\) reduction before stopping. Kimi performs a related set of checks through low-order recovery examples, including the two-site and three-site cases.

\paragraph{A075: AdS scalar bulk-to-boundary convolution.}

The task requires connecting the bulk solution, source normalization, distributional boundary limit, and CFT scaling law without conflating pointwise decay with source recovery.

\begin{tcolorbox}[
    trainingcase,
    colbacktitle=glmviolet!12,
    title={Kimi-K2.7-Code}
]
\textcolor{glmviolet}{\textbf{Final-answer fragment}}

\vspace{1.5mm}

{\color{textgray}\normalfont
With the source normalization fixed,
\[
\lim_{x_0\to0}
x_0^{-(d-\Delta)}
\Phi(x_0,\mathbf x)
=
\phi_0(\mathbf x).
\]
Away from the source, the leading behavior is the faster \(x_0^\Delta\) tail. The near-boundary expansion is
\[
\Phi
=
x_0^{d-\Delta}\phi_0
+
x_0^\Delta\psi
+\cdots,
\qquad
\psi(\mathbf x)
=
C_\Delta
\int d^d\mathbf x'
\frac{\phi_0(\mathbf x')}
{|\mathbf x-\mathbf x'|^{2\Delta}}.
\]
The coefficient \(\psi\) is the linear response, so
\[
\langle
\mathcal O(\mathbf x)
\mathcal O(\mathbf x')
\rangle
\propto
|\mathbf x-\mathbf x'|^{-2\Delta}.
\]
}
\end{tcolorbox}

\begin{tcolorbox}[
    trainingcase,
    colbacktitle=red!10,
    title={Qwen3.6-35B-A3B Base}
]
\textcolor{red!65!black}{\textbf{Final-answer fragment}}

\vspace{1.5mm}

{\color{textgray}\normalfont
Set \(\mathbf y=x_0\mathbf z\) in the boundary convolution. Then
\[
\phi(x_0,\mathbf x)
=
\mathcal N_\Delta
\int_{\mathbb R^d}
\frac{d^d\mathbf z}
{(1+|\mathbf z|^2)^\Delta}
\phi_0(\mathbf x+x_0\mathbf z).
\]
Since
\[
\int_{\mathbb R^d}
\frac{d^d\mathbf z}
{(1+|\mathbf z|^2)^\Delta}
=
\frac{\pi^{d/2}\Gamma(\Delta-d/2)}
{\Gamma(\Delta)},
\]
the Dirichlet condition fixes
\[
\mathcal N_\Delta
=
\frac{\Gamma(\Delta)}
{\pi^{d/2}\Gamma(\Delta-d/2)}.
\]
Conformal invariance then fixes the boundary two-point power to \(2\Delta\).
}
\end{tcolorbox}

\begin{tcolorbox}[
    trainingcase,
    colbacktitle=qwenblue!20,
    title={Qwen3.6-35B-A3B + SFT}
]
\textcolor{qwenblue}{\textbf{Final-answer fragment}}

\vspace{1.5mm}

{\color{textgray}\normalfont
The kernel vanishes pointwise for \(\mathbf x\neq\mathbf x'\) but is singular at the source. With
\[
C_\Delta
=
\frac{\Gamma(\Delta)}
{\pi^{d/2}\Gamma(\Delta-d/2)},
\]
its rescaled distributional limit is
\[
\lim_{x_0\to0}
x_0^{\Delta-d}
K_\Delta(x_0,\mathbf x-\mathbf x')
=
\delta^d(\mathbf x-\mathbf x').
\]
The dual boundary dimension and two-point function are
\[
\Delta
=
\frac d2+\sqrt{\frac{d^2}{4}+m^2},
\qquad
\langle O(\mathbf x)O(\mathbf x')\rangle
=
\frac{C_{OO}}
{|\mathbf x-\mathbf x'|^{2\Delta}}.
\]
}
\end{tcolorbox}

\textbf{Observed pattern.} The main distinction in this example is the treatment of the boundary limit. The SFT response explicitly separates pointwise decay away from the source from distributional recovery of the boundary source. The base response instead concentrates on the change of variables that fixes the normalization constant, while Kimi connects the normalized bulk solution directly to the normalizable coefficient and boundary response.

\paragraph{Cross-case pattern.}

Across the three prompts, the recurring SFT-associated difference lies primarily in the organization of the derivation rather than in a uniform change of the mathematical endpoint. The SFT responses more consistently expose normalization choices, boundary conditions, limiting cases, and local consistency checks as explicit steps before termination. The base student generally follows a shorter prompt-aligned progression, while Kimi often organizes the answer around the central theoretical structure and uses selected structural or low-order checks. These examples are qualitative, but they show a consistent separation between deriving the main result and checking the conditions under which that result is valid.

\paragraph{Relation to OPD degradation.}

The direct-OPD diagnostics in Appendix~\ref{app:opd-degeneration} and the SFT comparisons here address distinct questions. The former document repetitive continuations and premature termination during training; the latter show improved benchmark performance and illustrate the organization of derivations in SFT responses. Together with the main results, these observations motivate using SFT as an initialization stage and show that subsequent OPD with ESCD can provide further gains. They do not, however, establish that SFT initialization prevents the observed degradation; assessing this effect requires matched OPD runs with and without SFT initialization under otherwise identical training and evaluation conditions.

\subsection{Summary}
\label{app:escd-analysis-summary}

These analyses clarify the mechanisms and scope of ESCD. The local gradient comparisons favor the event-set target over a single-completion target, while the coverage analysis shows that one additional student action can complete representative residual constraints accounting for over 99\% of the observed compatible teacher probability mass in the studied tokenizer pairs. Together, these results support the completion-set design and the practical availability of visited-child supervision. Additional OPD diagnostics characterize repetitive continuation and premature termination in the Kimi-to-Qwen configuration, while the SFT comparisons provide complementary evidence on benchmark performance and derivation structure.

\end{document}